\pdfoutput=1

\documentclass[11pt]{article}

\usepackage[table]{xcolor}%
\usepackage{graphicx} %
\usepackage{tabularx}
\usepackage{booktabs}
\usepackage{amsmath}
\usepackage{hyperref}
\usepackage{amsfonts}
\usepackage{multirow}
\usepackage{xurl}

\usepackage{algorithm}
\usepackage{algpseudocode}

\definecolor{applegreen}{rgb}{0.0, 0.5, 0.0}

\usepackage[]{acl}

\usepackage{times}
\usepackage{latexsym}

\usepackage{array}
\newcolumntype{?}{!{\vrule width 2pt}}

\usepackage[T1]{fontenc}

\usepackage[utf8]{inputenc}

\usepackage{microtype}
\usepackage{verbatim}
\usepackage{fontawesome5}

\usepackage{inconsolata}

\newcommand{\projectname}{\mbox{CaLF}}
\newcommand{\secref}[1]{\S\ref{#1}}

\title{Learning to Generate Answers with Citations via \\Factual Consistency Models}

\author{Rami Aly$^1$\thanks{\hspace{0.3em}Work done while interning at Amazon Web Services.}, Zhiqiang Tang$^2$, Samson Tan$^2$, George Karypis$^2$\\
      \textsuperscript{1}University of Cambridge \hspace{1em}
      \textsuperscript{2}Amazon Web Services \hspace{1em} \\
      \texttt{rami.aly@cl.cam.ac.uk}, \texttt{\{zqtang|samson|karypis\}@amazon.com}
    }

\begin{document}
\maketitle
\begin{abstract}
Large Language Models (LLMs) frequently hallucinate, impeding their reliability in mission-critical situations. One approach to address this issue is to provide citations to relevant sources alongside generated content, enhancing the verifiability of generations. However, citing passages accurately in answers remains a substantial challenge. This paper proposes a weakly-supervised fine-tuning method leveraging factual consistency models (FCMs). Our approach alternates between generating texts with citations and supervised fine-tuning with FCM-filtered citation data. Focused learning is integrated into the objective, directing the fine-tuning process to emphasise the factual unit tokens, as measured by an FCM. Results on the ALCE few-shot citation benchmark with various instruction-tuned LLMs demonstrate superior performance compared to in-context learning, vanilla supervised fine-tuning, and state-of-the-art methods,  with an average improvement of $34.1$, $15.5$, and $10.5$ citation F$_1$ points, respectively. Moreover, in a domain transfer setting we show that the obtained citation generation ability robustly transfers to unseen datasets. Notably, our citation improvements contribute to the lowest factual error rate across baselines.

\end{abstract}

\section{Introduction}

Large language models \citep[LLMs;][]{brown2020} have showcased remarkable capabilities in information-seeking tasks. However, concerns persist regarding their trustworthiness, stemming from their opaque nature and factual hallucinations in their responses \citep{zhang2023siren, mallen-etal-2023-trust}. One proposed solution %
involves the production of citations to retrieved text passages alongside LLM-produced statements \cite{liu-etal-2023-evaluating, rashkin2023measuring}. By integrating citations into generated responses, readers can easily verify LLMs statements. The ability to accurately produce citations enables LLMs to generate responses more closely aligned with cited sources, alleviating hallucinations \citep{gao-etal-2023-enabling, yue-etal-2023-automatic}.

   \begin{figure}
            \centering
            \small
            \fbox{\begin{minipage}{23em}
            \textbf{Question}: Who has the highest goals in world football? \\
            
            \textbf{Retrieved Passages}: 
            
            \textbf{[1]} \ldots \textcolor{red}{Pelé's 1281 goals are recognized by FIFA as the highest total} achieved by a professional footballer, although the Soccer Statistic Foundation (rssf) \textcolor{blue}{recognizes only 767 goals} in official mode, \textcolor{blue}{occupying the third place after Josef Bican (805) and Romario (772)} \ldots  \\
            \vspace{-1em}
            
            \textbf{[2]} have beaten Gerd Müller's record of 85 goals in a year, the Football Association of Zambia \ldots %

            \noindent\rule{23em}{0.4pt}

            \textbf{Answer (LlaMA-2-Chat-70B)}: According to IFFHS International Federation of Football History and Statistics, \textcolor{red}{Lionel Messi} has the highest goals in world football with 1281 goals \textbf{[1]}. However, some sources, such as the Soccer Statistic Foundation (rssf), recognize only 767 goals in official mode, placing Messi in third place after Josef Bican (805) and Romario (772) \textbf{\textcolor{blue}{[2]}}. %

            \end{minipage}}
        \caption{An ALCE-ASQA question with a generated answer prompted via in-context learning. Two error classes are common: %
        information not supported by the sources (\textcolor{red}{red}) and incorrect citation to the sources (\textcolor{blue}{blue}).} %
        \label{fig:example}
    \end{figure}

Despite its significance, accurate citation generation proves to be challenging. State-of-the-art LLMs, such as ChatGPT \citep{openai2023gpt}, and commercial generative chat engines, such as Bing Chat, produce accurate citations only for less than 60\% of generated statements \citep{gao-etal-2023-enabling, liu-etal-2023-evaluating}. Figure \ref{fig:example} illustrates typical citation errors, including hallucinated statements or citations associated with incorrect claims. Hence, there is a necessity to train LLMs to generate citations accurately. This paper focuses on teaching LLMs to generate citations for retrieval-augmented long-form question answering (LFQA), tackling two main challenges: the scarcity of high-quality labeled data at scale and the risk of compromising original language and generalization capacities during fine-tuning for citation generation.

\begin{figure}
	\centering
	\includegraphics[width=1\linewidth, trim={0cm 0cm 0cm 0cm},clip]{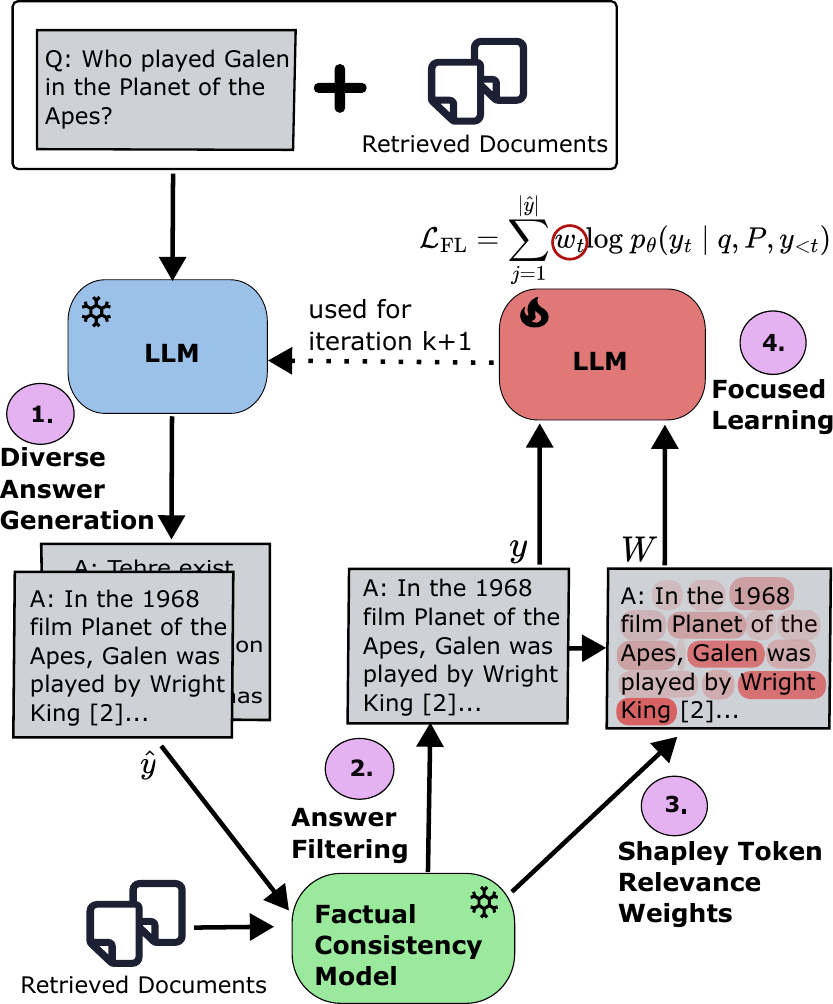}
	\caption{A schematic view of our iterative citation fine-tuning method \projectname. It uses a factual consistency model to: i) create weakly supervised training instances by filtering diversely sampled responses, ii) adjust the loss contribution of each answer token according to its Shapley relevance for factual consistency prediction.}
	\label{fig:method}
\end{figure}

To address these challenges, we present \textbf{\projectname} \ (\underline{\textbf{C}}it\underline{\textbf{a}}tion \underline{\textbf{L}}earning via \underline{\textbf{F}}actual Consistency Models), a fine-tuning strategy that enables LLMs to learn citation generation without sacrificing their language capacities. %
As illustrated in Figure \ref{fig:method}, the cornerstone of our approach is factual consistency models \citep[FCMs;][\textit{inter alia}]{kryscinski-etal-2020-evaluating}, initially introduced as a neural measure of consistency between a claim and its context. We use FCMs to gauge whether cited passages support a generated statement. Our method incorporates FCMs in two designs. Firstly, we propose a weakly-supervised training strategy, where an LLM generates diverse responses with citations, an FCM filters high-quality citation data, and the LLM is fine-tuned on the filtered data. Secondly, we utilize focused learning to adjust the loss contribution of each answer token based on its factual relevance, as measured by an FCM. The intuition is to have the LLM concentrate on tokens related to factual knowledge during fine-tuning, minimizing the impact on its original language capacities.

We evaluate \projectname \ on various LLMs, including Llama2 \citep{touvron2023llama}, Mistral-Instruct, and MistralOrca \citep{jiang2023mistral}. On the ALCE automatic few-shot citation evaluation benchmark \citep{gao-etal-2023-enabling}, \projectname \  enhances citation metrics over the in-context learning and baseline fine-tuning, with an average improvement of $34.1$ and $15.5$ F$_{1}$, respectively, while maintaining fluency and correctness. All LLMs trained via \projectname \ outperform the state-of-the-art model Self-Rag \citep{asai2023self} and ChatGPT \citep{openai2023gpt}, %
with an average improvement of $24.8$ and $10.5$ citation F$_{1}$ points, respectively. Domain transfer experiments, testing citation quality on a dataset different from the training dataset, highlight \projectname's ability to generalize across tasks and domains. Additionally, on the FactScore biography generation benchmark \citep{min-etal-2023-factscore}, \projectname \ demonstrates an improvement in factuality. Finally, human evaluation results indicate that \projectname \ yields more preferable answers compared to the fine-tuning baseline.\footnote{Code available at \url{https://github.com/amazon-science/learning-to-generate-answers-with-citations}}

\section{Related Work}

\paragraph{LFQA with Citations} To produce citations alongside a response, the generation can be conditioned on a few high-quality in-context examples with embedded citations \citep{gao-etal-2023-enabling, li2023helma}. %
In contrast, \citet{gao-etal-2023-rarr, bohnet2022attribution} propose to edit an already generated response post-hoc to attribute to retrieved sources, causing computational overhead during inference. %
Alternatively, proprietary work has explored human preference learning for citation production \citep{menick2022teaching, laMDA, nakano2021webgpt}. %
Reinforcement learning from human feedback is expensive and typically more brittle than supervised fine-tuning. Very recently, 
\citet{asai2023self} incorporate critique tokens, which serve as a feedback and citation mechanism, into GPT-4 obtained instruction-tuning data. These tokens allow for flexible retrieval-augmented generation with citations. %
In contrast to previous work, \projectname \ incorporates openly available FCMs as critique models into the fine-tuning process of already instruction-tuned LLMs while not modifying inference, maintaining efficiency.

\paragraph{Factual Consistency Models}

FCMs assess whether all the factual information in a claim is consistent w.r.t to the information conveyed in its grounding text.  The task shares strong similarities to natural language inference (NLI) \citep{bowman-etal-2015-large, dagan2005pascal}. However, in contrast to NLI, factual consistency is not evaluated on subjective or opinionated statements. %
Work that explores FCMs for \emph{improving} the generation is mainly constrained to summarization. \citet{aharoni-etal-2023-multilingual} filter samples from a large-scale dataset according to an FCM while \citet{muller-etal-2023-evaluating} use an FCM to rerank and select source passages for cross-lingual summarization. \citet{tian2023fine} incorporate FCMs to improve factuality %
via direct preference optimization \citep{rafailov2023direct}. Instead of using FCMs, \citet{deng-etal-2023-towards} improve the factuality of generations by measuring the cosine similarity between a token's embedding and relevant knowledge to adjust a token's loss contribution.
Our work is the first to explore the use of model explainability mechanisms, namely Shapley values \citep{shapley1953value}, for incorporating FCMs directly into the fine-tuning process of an LLM. %

\section{Preliminaries}
    
\paragraph{Task description.} Given an information-seeking question $q$, such as shown in Figure \ref{fig:example}, the task is to generate a long-form answer $\hat{y}=\{s_1, \cdots, s_n\}$, consisting of sentences $s_i$, conditioned on passages $P$ retrieved from a knowledge base. 
A long-form answer \emph{with citations} needs to ensure that one or multiple relevant passages $C_i \subseteq P$ are cited in each generated sentence $s_i$ (indicated by brackets with a passage index, e.g. ``[1]'' for $p_1$), such that the generated information in $s_i$ follows from the cited passages $C_i$.
This task definition strictly requires all facts to originate from the retrieved passages, ensuring that $\hat{y}$ is fully verifiable by $P$. We further assume the availability of few-shot training samples $(q, P, y) \in \mathcal{D}$ to learn citation production. %

\paragraph{Generation with LLMs.} In this work, we are interested in using LLMs to generate long-form answers with citations. An answer $\hat{y}$ is generated autoregressively, computing the next token distributions conditioned on the question $q$, retrieved passages $P$, and the answer generated so far $\hat{y}_{<t}$: $\prod_{t=1}^{\mid \hat{y} \mid} p_{\theta}(\hat{y}_t \mid q, P, \hat{y}_{<t})$,
with $\theta$ being the parameterization of the LLM. 
Consequently, a model is updated on a gold answer $y$ by minimizing the negative log-likelihood loss (NLL):
\begin{align}
\mathcal{L}_{\text{NLL}} = -\frac{1}{\mid y \mid}\sum_{t=1}^{\mid y \mid} \text{log } p_{\theta}(y_t \mid q, P, y_{<t}),
\label{Eq:nll}
\end{align}

\paragraph{Factual Consistency Models.} An FCM $\phi$ measures the factual consistency between a sentence $s_i$ and a collection of citations $C_i$:  $o = \phi(s_i, C_i)$,
with $o \in \{0, 1\}$  being the binary consistency prediction of the FCM. While FCMs are often modelled via modified NLI models that output a binary prediction directly \citep{gekhman-etal-2023-trueteacher, utama-etal-2022-falsesum}, %
AlignScore \citep{zha-etal-2023-alignscore} produces a single calibrated continuous value $o= p_{\phi}(\text{consistent} \mid s_i, C_i) \in [0,1]$. 
In our scenario, such a scalar is beneficial for computing Shapley values and for controlling the consistency's strictness. We refer to  \citet{mosca-etal-2022-shap} for background on Shapley values and the SHAP framework in the context of NLP.%

\section{Citation Learning via FCMs}

\projectname \ is a fine-tuning strategy for producing long-form answers with citations. %
\projectname's \ main assumption is that an FCM $\phi$ can be leveraged as a supervision signal to improve citation quality of an LLM $\mathcal{F}$ via an iterative training procedure, as illustrated in Figure \ref{fig:method} and Algorithm \ref{alg:method}. \projectname \ alternates between two modes in each iteration $k$. First, it generates weakly-supervised training data $\mathcal{\hat{D}}_{k}$ to enrich the training corpus by sampling diverse answers from a fine-tuned LLM $\mathcal{F}_{k-1}$ and filters them via the FCM $\phi$ (Sec. \ref{sec:attribution-training}). Second, it fine-tunes the LLM $\mathcal{F}_{k-1}$ on $\mathcal{\hat{D}}_{k} + \mathcal{D}$  with a modification of the NLL objective, which re-weights the loss contribution of individual tokens according to their importance for ensuring factual consistency, as measured by the FCM $\phi$ (Sec.~\ref{sec:focused-learning}). %
The number of iterations is determined dynamically by stopping when the proportion of filtered examples over the candidates %
does not improve between iterations or once the maximum number of iterations is reached. %

\begin{algorithm}%
\small
\caption{The training procedure of \projectname.}
\label{alg:method}
\begin{algorithmic}[1]
\Require LLM $\mathcal{F}$; FCM $\phi$; Retriever $\mathcal{R}$; questions $\mathcal{X}$ and answer facts $\mathcal{A}$; few-shot examples $\mathcal{D}$; Iterations $K$.
\Ensure Fine-tuned LLM $\mathcal{F}_K$; Citation Data $\mathcal{\hat{D}}_K.$
\State $\mathcal{U}_0 \gets  \{\oplus_{s_i \in y}(\text{Norm}(\text{SHAP}_{\phi}(s_i)))$ 
 $\mid (x, P, y) \in \mathcal{D} \}$
\State $\mathcal{W}_{0} \gets $ \{Align$(U, \mathcal{T}_{\phi}(\hat{y}), \mathcal{T}_{\mathcal{F}}(\hat{y}) \mid U \in \mathcal{U}_0\}$ 
\State $\mathcal{P} \gets \{ \mathcal{R}(x, \text{KB}) \mid x \in \mathcal{X} \}$
\State $k \gets 0$
\While{$k \leq K$ \textbf{and} $\frac{|\hat{D}_{k-1}|}{|\hat{Y}_{k-1}|} \geq |\frac{\hat{D}_{k-2}}{\hat{Y}_{k-1}}|$}
\If{$k=0$}  \Comment{\textcolor{red}{Data Generation} (\secref{sec:attribution-training})}
\State $\hat{Y}_k \gets$ Diverse Sampl.$_{\text{IC}}(\mathcal{F}, \mathcal{X}, P, \mathcal{D}$)
\Else{} %
\State $\hat{Y}_k \gets$ Diverse Sampl.$(\mathcal{F}_{k-1}, \mathcal{X}, P$)
\EndIf
\State $\hat{D}_k \gets$ \{$(x, P, \hat{y}) \mid \hat{y} \in \hat{Y}_k \land \mathcal{Q}(\hat{y}, A, P) = 1$\} \hspace{-0.3em}
\State $\mathcal{U}_k \gets  \{\underset{s_i \in \hat{y}}{\oplus}\text{Norm}(\text{SHAP}_{\phi}(s_i))) \mid (x, P, \hat{y}) \in \hat{\mathcal{D}}_k \}$ 
\State $\mathcal{W}_{k} \gets $ \{Align$(\mathcal{T}_{\phi}(\hat{y}), \mathcal{T}_{\mathcal{F}}(\hat{y}) \mid U \in \mathcal{U}_k\}$ 
\State $\mathcal{F}_k \gets $ Update $\mathcal{F}$ via FL  \Comment{\textcolor{red}{Focused Learning} (\secref{sec:focused-learning})} \\ 
\hspace{8.5em} $\nabla \mathcal{L}_{\text{FL}}(\mathcal{\mathcal{D} + \hat{D}}_{k}, \mathcal{W}_{0} + \mathcal{W}_{k})$ 
\State $ k \gets k + 1$
\EndWhile
\end{algorithmic}
\end{algorithm}

\subsection{Answer Generation for Training}
\label{sec:attribution-training}
To generate weakly supervised training data $(x, P, \hat{y}) \in \mathcal{\hat{D}}_k$ with answer $\hat{y}$ containing citations to passages $P$, we assume the availability of a collection of information-seeking questions $x \in \mathcal{X}$ and a list of atomic facts $A$ expected in an answer
to $x$.  %
For questions $\mathcal{X}$, an LLM generates a collection of answer candidates $\mathcal{\hat{Y}}$, conditioned on retrieved passages $P$, selected by an out-of-the-box retrieval system $\mathcal{R}$. As indicated in Algorithm \ref{alg:method}, we produce answer candidates using either the fine-tuned model $F_{k-1}$ from the previous iteration, or, in the case $k=0$, we use in-context prompting with the few-shot examples $\mathcal{D}$. We focus on the generation of \emph{diverse answer candidates} to enrich the weakly supervised training data. First, we use sampling strategies such as nucleus sampling \citep{Holtzman2020The}, temperature scaling \citep{guo2017calibration}, and diverse beam search \citep{diverse-decoding-2018}. Second, we consider citation replacements %
in answer sentences $s_i$ in $\hat{y}$ to diversify the answer candidates beyond the output of the LLM. Specifically, for generated citations $C_{i}$ in a sentence $s_i$, we generate two citation replacements, sampled according to the passage probability measured via the retriever $\mathcal{R}$ since the direct computation of $\phi(s_i, C_i)$ over all citation options is infeasible.

Each answer candidate $\hat{y} \in \hat{\mathcal{Y}}$ is subsequently filtered via an answer quality assurance function $\mathcal{Q}_{\phi}: \hat{y} \rightarrow \{0, 1\}$, measured via the FCM $\phi$, to obtain $\hat{\mathcal{D}_t}$ with weakly-supervised cited answers $\hat{y}$:

\begin{align*}
\mathcal{Q} = \begin{cases}
        1  & \text{if} \hspace{1em} \text{Citation-Recall}(\hat{y}, C, \phi) > \Theta \\
        & \hspace{0.7em} \land \text{ Citation-Precision}(\hat{y}, C, \phi)  > \Theta \\
        & \hspace{0.7em} \land \text{ Correctness}(\hat{y}, A, \phi)  > \Theta \\
        0  & \text{else},
        \end{cases}
\end{align*}
with $C$ being the citations assigned to sentences in $\hat{y}$, and $\Theta$ being a dynamically determined quality threshold that adjusts such that the size $\hat{D}_t$ is above a minimum viable size.  %
\text{Citation-Recall}$(\hat{y}, C)=\frac{1}{n}\sum_{s_i\in \hat{y}} \phi(s_i, C_i)$ measures the factual coverage of citations, \text{Citation-Precision}$(\hat{y}, C)=\frac{1}{|C|} \sum_{C_i \in C} \frac{1}{|C_i|}\sum_{c_{i, j}\in C_i}  \text{max}(\phi(s_i, c_{i,j}), 1-\phi(C_i \text{\textbackslash} \{c_{i,j\}, s_i}))$ measures the relevance of citations, and Correctness$(\hat{y}, A)$ measures the proportion of facts $A$ covered in $\hat{y}$. These definitions largely align with the ones in the benchmark of \citet{gao-etal-2023-enabling} for evaluating citations.

\begin{figure}[ht]
	\centering
	\includegraphics[width=1\linewidth, trim={0cm 0cm 0cm 0cm},clip]{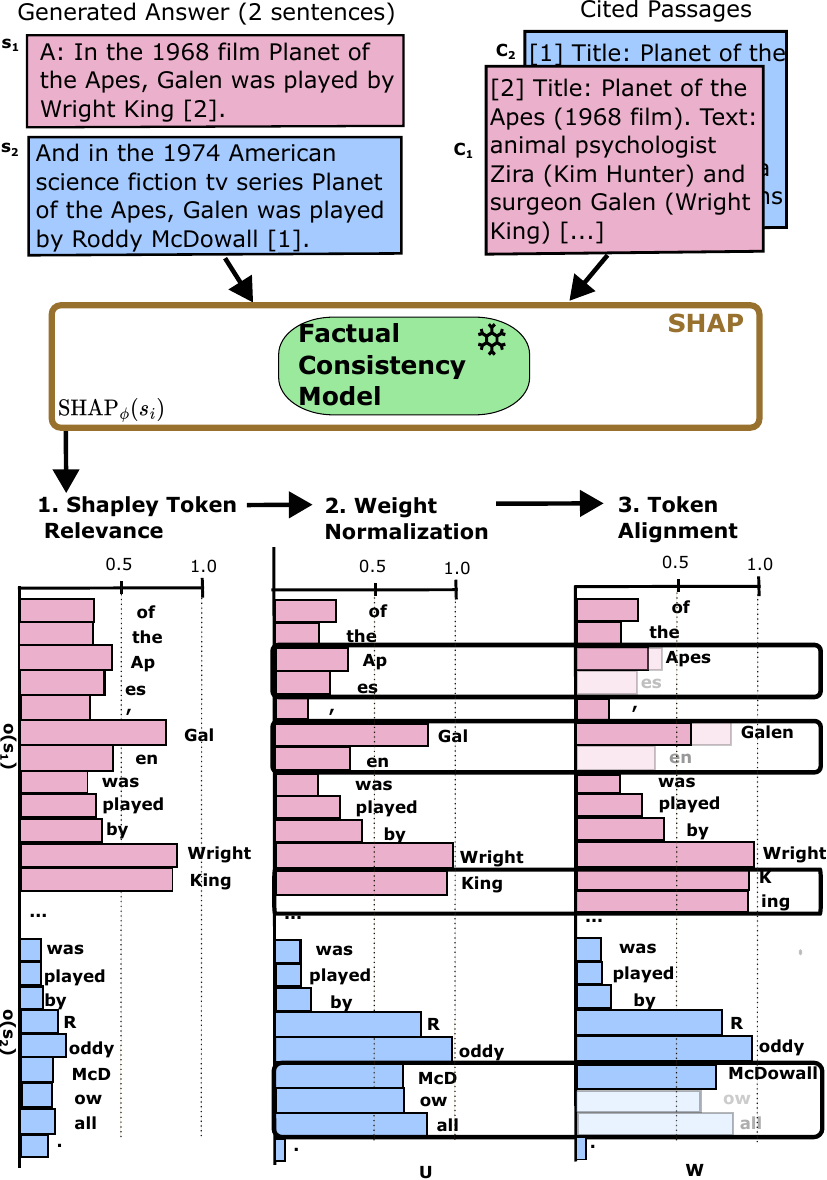}
	\caption{The computation of relevance weights $W$ for rescaling the loss according to Eq.~\ref{Eq:fl}. We first use SHAP to measure the token importance for predicting $\phi(s_i, C_i)= o_{i} $. We adjust for differences in scale of $W_{\phi,i}$ for sentences $s_i$ and differences in tokenization between the FCM and the LLM.} %
	\label{fig:method-focused-learning}
\end{figure}

\subsection{Focused Learning for Factual Consistency}
\label{sec:focused-learning}
To emphasize the learning of producing citations, we measure the relevance for each token in an answer %
for ensuring the factual consistency between $y$ and the retrieved passages $P$. We subsequently modify the NLL loss computation (see Eq.~\ref{Eq:nll}) for the instruction-tuning of the LLM $\mathcal{F}$ by re-weighting the loss contribution of the $t-$th token according to relevance weights $w_t \in W$:

\begin{align}
\mathcal{L}_{\text{FL}} = -\frac{1}{\mid y \mid}\sum_{j=1}^{\mid y \mid} w_t \text{log } p_{\theta}(y_t \mid q, P, y_{<t}).
\label{Eq:fl}
\end{align}

In contrast to NLL where the loss is computed as the arithmetic mean over the token-level loss, our focused learning loss $\mathcal{L}_{\text{FL}}$ emphasizes tokens which are considered of higher importance for ensuring factual consistency between a generated statement and the cited passages according to an FCM $\phi$.

The mechanism for obtaining the relevance weights $W$ is illustrated in Figure \ref{fig:method-focused-learning}. We first compute and normalize Shapley values for the factual consistency prediction between $y$ and its cited passages using the FCM $\phi$, obtaining $U$. These normalized relevance values for the tokens of the FCM are subsequently mapped to tokens of the LLM $\mathcal{F}$ via an alignment algorithm. 
As seen in Figure \ref{fig:method-focused-learning}, given a generated statement such as ``\emph{In the 1968 film Planet of the Apes, Galen was played by Wright King.}'', the relevance weights $W$ consider factual tokens such as \emph{Wright} and \emph{King} more important than \emph{of} and \emph{was}, emphasizing units of information important for factually accurate citation.

\paragraph{Computation of Token Relevance.}
We first compute relevance weights $U$ over the FCM $\phi$ by computing Shapley values for the factual consistency prediction between an answer's sentence $s_i$ and its citations $C_i$: $o_i=\phi(s_i, C_i)$. Shapley values assign importance scores $w_t$ to each feature (here token) in $s_i$ concerning prediction $o_i$. %
Since Shapley values distribute the prediction score $o_i$ along all the tokens of sentence $s_i$, the value of $o_i$ and the length of $s_i$ impacts the scale of assigned values, as shown in Figure \ref{fig:method-focused-learning}, potentially biasing the loss towards shorter sentences and amplifying idiosyncrasies of the FCM.   
Thus, we normalize the assigned Shapley values for each sentence via min-max normalization Norm($U_{i}$): \{$\frac{u_{t} - \text{min}(U_{i})}{\text{max}(U_{i}) - \text{min}(U_{i})} \mid u_t \in U_{i} \}$, with $U_{i}$ being token weights for sentence $s_i$. Thus, the token with the highest and lowest Shapley value is assigned a relevance score of $0$ and $1$, respectively.\footnote{The exploration of alternative normalization functions is left to future work.} The computation of weights $U$ for a given response $y$ can subsequently be summarized as:
 \begin{align}
U = \oplus_{i=1}^{n}\text{Norm}(\text{SHAP}_{\phi}(s_i))  %
 \end{align}
with $\oplus$ being a concatenation operator. Since citation tokens do not bear any semantic meaning themselves, they are excluded from the computation of $\phi(s_i, C_i)$ and are thus not yet contained in $U$. Therefore, we further insert a weight of $1$ for each citation token of $y$ into $U$.

\begin{table*}[ht!]
	\centering
	\resizebox{1\linewidth}{!}{
	\begin{tabular}{l | r r r r r | r  r r r r}
            \toprule
            Method  & \multicolumn{5}{c|}{\cellcolor{gray!25} ALCE-ASQA } & \multicolumn{5}{|c}{\cellcolor{gray!25} ALCE-ELI5 }  \\
             \hline
              & Similarity & Fluency & Correct. &\textbf{Correct.} &  \textbf{Citation} &  Similarity & Fluency & Correct. & \textbf{Correct.} &  \textbf{Citation} \\
              & Rouge-L & MAUVE & EM Rec. & \textbf{in $P$} & \textbf{F$_1$} &Rouge-L & MAUVE & EM Rec. & \textbf{in $P$} & \textbf{F$_1$}  \\
            \hline
            ChatGPT \citep{gao-etal-2023-enabling} & -- & 	66.6 & 40.4 & -- &  	73.1 & -- & 	57.2 & 12.0 & -- &  50.5 \\
            GPT-4 \citep{gao-etal-2023-enabling} & -- & 	67.1 & 41.3 & -- &  	 71.9 & -- & 	38.4 & 14.2 & -- &  	46.9 \\
            AGREE \citep{ye2024effective} & -- &-- & 40.9  & -- & 75.1 & -- & -- & -- & -- & --\\
            \hline
            Self-RAG 7B \citep{asai2023self} & 35.7&	74.3 & 30.0 & -- &	67.3 & 16.9 & 32.6 & 9.7 & 5.4 & 27.6\\
            BP, T5-3B \citep{fierro2024learning} & -- &	-- & 33.8 & -- &	77.8 & -- & -- & 5.2 & -- & 60.9\\
            \hline
            \textbf{Llamav2-7B-chat} \hspace{0.8em} In-context & 35.9$_{0.3}$  & 77.8$_{3.1}$ & 35.0$_{0.6}$ & 25.7$_{0.6}$ & 49.9$_{1.0}$ & 20.5$_{0.2}$ & 36.2$_{2.5}$ & 17.7$_{0.6}$ & 10.8$_{0.6}$  &  38.2$_{0.6}$  \\
            \multicolumn{1}{r|}{Few-shot FT} & 34.9$_{  0.4}$ & 69.2$_{  4.3}$ & 32.0$_{  0.4}$ & 22.3$_{  0.7}$ & 55.0$_{  1.8}$ & 21.3$_{  0.2}$ & 58.2$_{  2.2}$ & 17.8$_{  0.6}$ & 11.2$_{  1.1}$ & 48.7$_{  2.9}$ \\
           \multicolumn{1}{r|}{Ours} & 37.8$_{0.4 }$ & 86.0$_{3.7 }$ & 37.7$_{0.6 }$ & 29.3$_{0.4 }$ & 70.4$_{2.5 }$  & 20.8$_{  1.0}$ & 59.6$_{  11.5}$ & 17.0$_{  0.3}$ & 11.9$_{  0.2}$ & 66.5$_{  5.9}$\\
            \hline
            \textbf{Mistral-Instr. 7B } \hspace{0.8em} In-context & 36.7$_{0.3}$ & 85.5$_{2.7}$ & 34.4$_{0.4}$ & 27.8$_{0.4}$ & 22.3$_{0.9}$ & 21.6$_{0.9}$ & 43.8$_{4.8}$ & 19.1$_{0.4}$ & 11.1$_{0.2}$ & 19.5$_{0.6}$ \\
            \multicolumn{1}{r|}{Few-shot FT} &  38.1$_{  0.2}$ & \textbf{87.7$_{  0.8}$} & 36.1$_{  0.9}$ & 29.4$_{  1.1}$ & 66.7$_{  4.5}$ & 20.5$_{  0.3}$ & 48.0$_{  4.9}$ & 15.5$_{  1.2}$ & 10.0$_{  0.6}$ & 49.9$_{  4.0}$ \\
            \multicolumn{1}{r|}{Ours} & 37.2$_{  1.2}$ & 84.5$_{  7.5}$ & 36.4$_{  1.6}$ & 30.0$_{  1.1}$ &  76.2$_{  1.9}$  & \textbf{21.8$_{  0.2}$} & 58.2$_{  4.0}$ & 19.5$_{  0.8}$ & 13.1$_{  0.2}$ &  66.0$_{4.7 }$ \\
            \hline
            \textbf{MistralOrca 7B} \hspace{1.2em} In-context &  38.7$_{0.1}$ & 54.7$_{1.8}$ & 40.2$_{0.3}$ & 31.9$_{0.2}$ & 55.6$_{0.8}$ & 20.9$_{0.1}$& 29.3$_{0.8}$ & \textbf{20.8$_{0.4}$} & 12.5$_{0.5}$ & 43.3$_{0.5}$ \\
            \multicolumn{1}{r|}{Few-shot FT}& 38.4$_{  1.8}$ & 78.6$_{  14.7}$ & $38.4_{  3.8}$ & 29.9$_{  4.7}$ & $62.6_{  3.6}$ & 19.4$_{ 1.9}$ & $60.5_{ 13.6}$ & $17.3_{ 1.8}$ & $10.9_{  1.2}$ & 57.7$_{ 6.5}$  \\
            \multicolumn{1}{r|}{Ours} & \textbf{40.3$_{ 0.2}$} & 84.0$_{ 3.3}$ & \textbf{41.7$_{ 1.2}$} & \textbf{34.5$_{ 0.5}$} & \textbf{81.5$_{  2.5}$} & 20.4$_{  1.5}$ & \textbf{62.7$_{ 4.6}$} & 18.4$_{ 2.1}$ & \textbf{13.1$_{ 0.7}$} & \textbf{73.1$_{ 4.2}$} \\
            \bottomrule
	\end{tabular}}
	\caption{Main results on ALCE using only $4$ training samples as $\mathcal{D}$, measured across $3$ random seeds. Fine-tuning via \projectname \ substantially improves citation quality (\textbf{Citation F}$_1$) and correct information grounded in passages (\textbf{Correct. in} $P$) over alternative training strategies and competitive baselines while maintaining factual coverage (\textbf{EM Recall}), fluency (\textbf{MAUVE}), and recall-oriented similarity (\textbf{ROUGE-L}) to gold responses without citations.}
        \label{tab:main-results}
\end{table*}

\paragraph{Feature Importance Mapping.}
Since the obtained relevance weights $U$ are specific to the tokenizer used for $\phi$, an LLM that uses a different tokenization may not directly apply $U$ to re-weight the loss $\mathcal{L}_{\text{FL}}$ (c.f. Eq.~\ref{Eq:fl}). To this end, we define an alignment function that maps the FCM token weights $U$ %
to LLM token weights $W$ %
for the same sequence $y$. %
The alignment function first maps the shortest possible token span $y_{\phi, l:l+m}$ from the FCM's tokenizer  $T_{\phi}$  to the span $y_{\mathcal{F},p:p+q}$ from the LLM's tokenizer $T_{\mathcal{F}}$, with $j,l \geq 1$.\footnote{Note that we implement this mapping function tailored to the specific tokenizers used by our models, see Appendix \ref{app:experimental-setup}.%
} The relevance score for each LLM token in $y_{\mathcal{F}, p:p+q}$ is computed as the average relevance score over the aligned FCM span $y_{\phi, l:l+m}$: $W_{p:p+q} = \{\frac{\sum_{l\leq t < m}(u_t)}{\mid y_{\phi, l:l+m} \mid} \mid  y_{t} \in y_{\mathcal{F}, p:p+q} \}$
In the example of Figure \ref{fig:method-focused-learning}, the weight for the LLM's token \emph{McDowall} is computed as the average over the weights for the three aligned FCM tokens (\emph{McD}, \emph{ow}, \emph{all}). Further, since both LLM tokens \emph{K} and \emph{ing} are aligned to the single token \emph{King} of the FCM, both tokens are set to the same relevance weight according to our algorithm.

\section{Evaluation}
\paragraph{Datasets \& Metrics.} We conduct experiments on long-form QA datasets of the ALCE citation benchmark \citep{gao-etal-2023-enabling}, namely their version of ASQA \citep{stelmakh-etal-2022-asqa} and ELI5 \citep{fan-etal-2019-eli5}. We further evaluate models on BIO \citep{min-etal-2023-factscore}. For our domain transfer experiments, we further consider Hagrid \citep{kamalloo2023hagrid} as a source for training. Since Hagrid's answers are generated by GPT-4 without annotations for factual coverage, we do not use it for evaluation as a target. For training, ALCE contains $\mid \mathcal{D} \mid =4$ 
cited gold instances. For Hagrid, we randomly sample $4$ instances for \projectname. %
Details are in Appendix \ref{app:datasets}.  %

We follow the official metrics and nomenclature of \citet{gao-etal-2023-enabling} to measure correctness (via EM Recall), fluency (via MAUVE \citep{pillutla2021mauve}), and citation F$_1$ (via an NLI-trained T5-11B model \citep{honovich-etal-2022-true-evaluating}). We further measure Rouge-L \citep{lin-2004-rouge} and introduce a more strict variation of their correctness metric: \emph{passage-grounded correctness} (Correct. in $\mathcal{P}$), which only considers information from responses which are supported by the retrieved passages $P$. %
 Subsequently, this metric ignores factual content produced by an LLM's parametric memory if not explicitly derivable from the retrieved passages, isolating factual grounding from a model's parametric memory. We use FactScore \citep{min-etal-2023-factscore} to evaluate the biographies of BIO. Detailed metric descriptions (e.g. citation recall and precision) and results are in Appendix \ref{app:evaluation}.  %

\paragraph{Experimental Setup.} 
Following recommendations for weakly supervised learning \citep{zhu-etal-2023-weaker} and few-shot learning \citep{alex2021raft}, we do not consider a validation set for hyperparameter-tuning, representing real-world scenarios more accurately. Due to computational constraints,  we use LoRA \citep{hu2022lora} for parameter-efficient fine-tuning of \projectname \ and our fine-tuning baselines. %
We use Alignscore as our FCM $\phi$ in all experiments unless otherwise mentioned. A threshold is used to map Alignscore's output $o$ into a binary prediction for $\mathcal{Q}$. Alignscore is substantially different from the FCM model used for evaluating citations, using a different architecture and training data (c.f. App.~\ref{app:experimental-setup}).
ASQA, Hagrid, and BIO use Wikipedia as their underlying knowledge base while ELI5 uses CommonCrawl. We use the same retrievers as  \citet{gao-etal-2023-enabling} and \citet{asai2023self} to maintain comparability. Further implementation details are in Appendix \ref{app:experimental-setup}.

\paragraph{Baselines.} 
We compare \projectname \ on identical instruction-tuned LLMs against both in-context prompting and few-shot fine-tuning (Few-shot FT) via Eq. \ref{Eq:nll}, trained on $\mathcal{D}$. In our domain transfer experiments, the entire $335$ training samples of Hagrid are used to train the fine-tuning baseline (FT). %
Furthermore, we consider state-of-the-art models, namely the most powerful in-context prompted baselines from \citet{gao-etal-2023-enabling}, ChatGPT (\texttt{gpt-3.5-turbo-0301}) and GPT-4 (\texttt{gpt-4-0613; 8K context window}). Due to context length limitations, in-context prompting uses $2$ randomly sampled instances. We further compare against the best results of AGREE \citep{ye2024effective} which use PaLM 2's text-bison-001. Finally, we evaluate against open-source models, including Self-RAG 7B \citep{asai2023self}, based on Llama2, and Blueprint (BP) \citep{fierro2024learning}, based on T5-3B. %

\begin{table*}[ht!]
	\centering
	\resizebox{1\linewidth}{!}{
	\begin{tabular}{l | r r r r r  ?  l | r r r r r }
            \toprule
              $\underset{\text{Source}\rightarrow \text{Target}}{\text{Method}}$ & Similarity & Fluency & Correct. & Correct. &  Citation & $\underset{\text{Source}\rightarrow \text{Target}}{\text{Method}}$  & Similarity & Fluency & Correct. & Correct. &  Citation \\
             &  RougeL & MAUVE & EM Rec. & in $P$ & F$_1$ &  & RougeL & MAUVE & EM Rec. & in $P$ & F$_1$  \\

            \hline
            Self-RAG 7B & 35.7 & 74.3 & 30.0 & -- &	67.3 & Self-RAG 7B & 16.9 & 32.6 & 9.7 & 5.4 & 27.6  \\
            \hline
             \multicolumn{12}{|c}{\cellcolor{gray!25} \textbf{Llama2-7B-chat} } \\
             $\underset{\rightarrow \text{ASQA}}{\text{Zero-Shot}}$ & 36.1 & 47.5 & 35.6 & 27.1 & 31.0 &  $\underset{\rightarrow \text{ELI5}}{\text{Zero-Shot}}$  & 20.0 & 26.5 & 15.3 & 9.7 & 24.4  \\
            $\underset{\text{ELI5}\rightarrow \text{ASQA}}{\text{Few-shot FT}}$ &  37.2 & 74.0 & 37.7 & 31.6 & 64.9 & $ \underset{\text{ASQA}\rightarrow \text{ELI5}}{\text{Few-shot FT}} $ & 17.1 & 20.8 & 11.4 & 6.8 & 32.9 \\
            $\underset{\text{ELI5}\rightarrow \text{ASQA}}{\text{Ours}}$ &37.1 & 75.7 & 36.1 & 30.4 & 73.1   & $\underset{\text{ASQA}\rightarrow \text{ELI5}}{\text{Ours}} $ &  \textbf{21.3} & 35.0 & 18.2 & 11.0 & 36.5 \\
            \hline
             \multicolumn{12}{|c}{\cellcolor{gray!25} \textbf{MistralOrca-7B} } \\
              $\underset{\rightarrow \text{ASQA}}{\text{Zero-Shot}}$ & 39.0 & 78.9 & 39.5 & 31.6 & 5.7 &  $\underset{\rightarrow \text{ELI5}}{\text{Zero-Shot}}$ & \textbf{21.3} & 35.0 & \textbf{22.2} & 12.6 &  10.4 \\
             $\underset{\text{ELI5}\rightarrow \text{ASQA}}{\text{Few-shot FT}}$ & 39.7 & \textbf{90.1} & 38.5 & 31.4 & 71.7 & $\underset{\text{ASQA}\rightarrow \text{ELI5}}{\text{Few-shot FT}}$  & 20.9 & 41.1 & 19.7 & 10.6 & 40.4 \\
             $\underset{\text{Hagrid}\rightarrow \text{ASQA}}{\text{Few-shot FT}}$  & 36.7 & 66.1 & 37.8 & 29.5 & 51.3 & $\underset{\text{Hagrid}\rightarrow \text{ELI5}}{\text{Few-shot FT}}$  & \textbf{21.3} & \textbf{58.1} & 20.9 & 12.7 & 32.3 \\
            $\underset{\text{ELI5}\rightarrow \text{ASQA}}{\text{Ours}}$ & \textbf{40.1} & 86.6 & \textbf{40.0} & \textbf{33.2} & 79.5 & $\underset{\text{ASQA}\rightarrow \text{ELI5}}{\text{Ours}}$  & 21.2 & 31.3 & 20.4 & 12.5 & \textbf{57.3}   \\
            $\underset{\text{Hagrid}\rightarrow \text{ASQA}}{\text{Ours}}$  & 39.7 & 80.8 & 38.6 & 32.3 &  \textbf{80.0}  & $\underset{\text{Hagrid}\rightarrow \text{ELI5}}{\text{Ours}}$   & 21.1 & 32.3 & 20.2 & \textbf{12.9} & 54.6 \\
            \bottomrule
	\end{tabular}}
	\caption{Results for our zero-shot domain transfer setting, when trained on a source dataset ($\mathcal{D}$) and evaluated on a different target dataset without any in-context instances. \projectname's citation quality %
 (\textbf{Citation F}$_1$) and passage-grounded correctness (\textbf{Correct. in $P$)} is superior to all baselines, without additional inference costs. %
 }
 \label{tab:domain-transfer}
\end{table*}

\subsection{Main Results}
Table \ref{tab:main-results} shows the main in-domain results, with mean and standard deviation computed over three seeds. \projectname \ improves citation F$_{1}$ across datasets and models by $34.1$ and
$15.5$ points over both in-context learning (In-context) and baseline fine-tuning (Few-shot FT), respectively.  
Impressively, \emph{all} tested LLMs with \projectname \ outperform Self-RAG, ChatGPT, and GPT-4
with an average improvement of $24.8$, $10.5$, and $12.9$ citation points, respectively. %
Moreover, \projectname \ achieves high citation scores while the overall quality of the response remains high. We observe modest improvement in correctness but substantial gains in grounded correctness, indicating that \projectname \ is better at including verifiable facts in its responses than our baselines. This contrasts observations for other models: GPT-4 trades off improvements in correctness at the cost of citation quality, resulting in ChatGPT having overall higher citation scores than GPT-4. Notably, \projectname \ improves on correctness over GPT-4 while also producing higher-quality citations than ChatGPT, beating both models at their best-performing metric. Similarly to GPT, BP produces accurate citations but has low correctness, especially on ELI5. Moreover, training via \projectname \ also leads to an overall improvement in ROUGE-L and MAUVE across models over the fine-tuning baseline by an average of $1.0$ and $5.5$, respectively.

\subsection{Domain Transfer}

We run domain transfer experiments to evaluate the generalization of \projectname's citation production, by training an LLM with \projectname \ on a source dataset and measuring performance on a different target dataset. %
Table \ref{tab:domain-transfer} shows our domain transfer results. In every source-target configuration and across instruction-tuned models, \projectname \ outperforms zero-shot in-context learning, Few-shot FT, FT, and Self-RAG in both citation quality and correctness. \projectname \ (using MistralOrca-7B) exhibits small variability regarding the training source $\mathcal{D}$, with a citation F$_{1}$ difference $\Delta_{*\rightarrow \text{ASQA}}$ of $0.5$ and  $\Delta_{*\rightarrow \text{ELI5}}$ of $2.7$, respectively. %
While \projectname's citation quality is comparable in-domain versus in a transfer setting on ASQA ($-\Delta 3.9$, see Table \ref{tab:main-results}), we observe larger differences on ELI5 ($-\Delta 11.9$), %
since for ELI5 the knowledge source and question scope greatly differ from the Wikipedia-based source datasets.

\subsection{Factuality}
\label{sec:factuality}
We evaluate \projectname's factual precision using FactScore. Results are shown in Table \ref{tab:factscore} for state-of-the-art methods taken from \citet{asai2023self}, our fine-tuning baseline, and \projectname \ with MistralOrca-7B. %
\projectname scores the highest with $83.4$, indicating that the improved citation quality translates to higher factual accuracy. Interestingly, while citation recall can be considered a stricter measure than the FactScore, the former is much higher for \projectname.  %
We postulate the difference is caused by retrieval inaccuracies.
Subsequently, we adjust predictions to abstain from answering if none of the passages' titles match with the BIO entity.\footnote{Abstaining is explicitly incorporated into FactScore.} As seen in Table \ref{tab:factscore}, the FactScore improves to $86.1$ and $88.9$ for our fine-tuning baseline and \projectname, respectively. 
While these results are promising they also highlight the importance of accurate retrieval and high-quality knowledge bases so that citation production can translate into improved factuality, an observation also made in \citet{menick2022teaching, kryscinski-etal-2020-evaluating}. %

\begin{table}[ht!]
	\centering
	\resizebox{1\linewidth}{!}{
	\begin{tabular}{l | r r}
            \toprule
            Method &  FS  & Citation Recall \\
            \hline
            ChatGPT w/o retrieval & 71.8 &--\\
            Llamav2-13B-chat  &  79.9 & -- \\
            Self-RAG 7B   &  81.2 & -- \\
            Self-RAG 13B &  80.2 & --\\
            \hline
            Few-shot FT & 78.7 & 69.3 \\
            Few-shot FT + Retrieval Filtering & 86.1 & 69.3 \\
            \hline
            Ours & 83.4 &  92.7  \\
            Ours + Retrieval Filtering & \textbf{88.9} & \textbf{92.7} \\
            \hline
	\end{tabular}}
	\caption{FactScore (FS) evaluation. MistralOrca7B is used with \emph{Few-shot FT} and \emph{Ours}. Other results are taken from \citet{asai2023self}. \textit{+ Retrieval Filtering}: the model abstains when no passage title matches the entity.}%
        \label{tab:factscore}
\end{table}

\section{Discussion}
\label{sec:discussion}

\paragraph{Ablation.}
Table \ref{tab:ablation} shows an ablation for the two mechanisms introduced in \projectname, the generation of weakly-supervised training data (WS) and the focused learning loss ($\mathcal{L}_{FL}$). By adding WS, we see an improvement of $18.3$ and $16.7$ citation recall and precision points on ASQA, respectively. By adding $\mathcal{L}_{FL}$, we further improve on top of WS by $6.3$ and $8.2$ recall and precision points, respectively. On ELI5, we observe similar improvements.

\begin{table}[ht!]
	\centering
	\resizebox{1\linewidth}{!}{
	\begin{tabular}{l | r r r}
            \toprule
             & Correctness &  Citation  & Citation \\
            Method & in $P$ & Recall & Precision \\
            \hline
            \multicolumn{4}{c}{\cellcolor{gray!25} ASQA } \\
            LLM & 25.8 & 57.5 & 55.2 \\
            LLM + WS & 29.0 (\textcolor{applegreen}{+3.2}) & 75.8 (\textcolor{applegreen}{+18.3}) & 71.9 (\textcolor{applegreen}{+16.7})\\
            LLM + WS + $\mathcal{L}_{FL}$ & \textbf{33.8} (\textcolor{applegreen}{+8.0}) & \textbf{84.1} (\textcolor{applegreen}{+24.6}) & \textbf{83.8} (\textcolor{applegreen}{+28.9})\\
            \multicolumn{4}{c}{\cellcolor{gray!25} ELI5 } \\
            LLM & 11.0 & 57.3 & 51.0 \\
            LLM + WS & 11.5 (\textcolor{applegreen}{+0.5}) & 61.5 (\textcolor{applegreen}{+4.2}) & 57.2 (\textcolor{applegreen}{+6.2}) \\
            LLM + WS + $\mathcal{L}_{FL}$ & \textbf{12.5} (\textcolor{applegreen}{+1.5}) & \textbf{72.1} (\textcolor{applegreen}{+14.8}) & \textbf{66.6} (\textcolor{applegreen}{+15.6}) \\    
            \hline
	\end{tabular}}
	\caption{Ablation for \projectname \ (MistralOrca-7B). WS: weakly-supervised training, $\mathcal{L}_{FL}$: Focused learning.}
        \label{tab:ablation}
\end{table}

\begin{figure}[ht]
	\centering
	\includegraphics[width=1\linewidth, trim={0cm 0cm 0cm 0cm},clip]{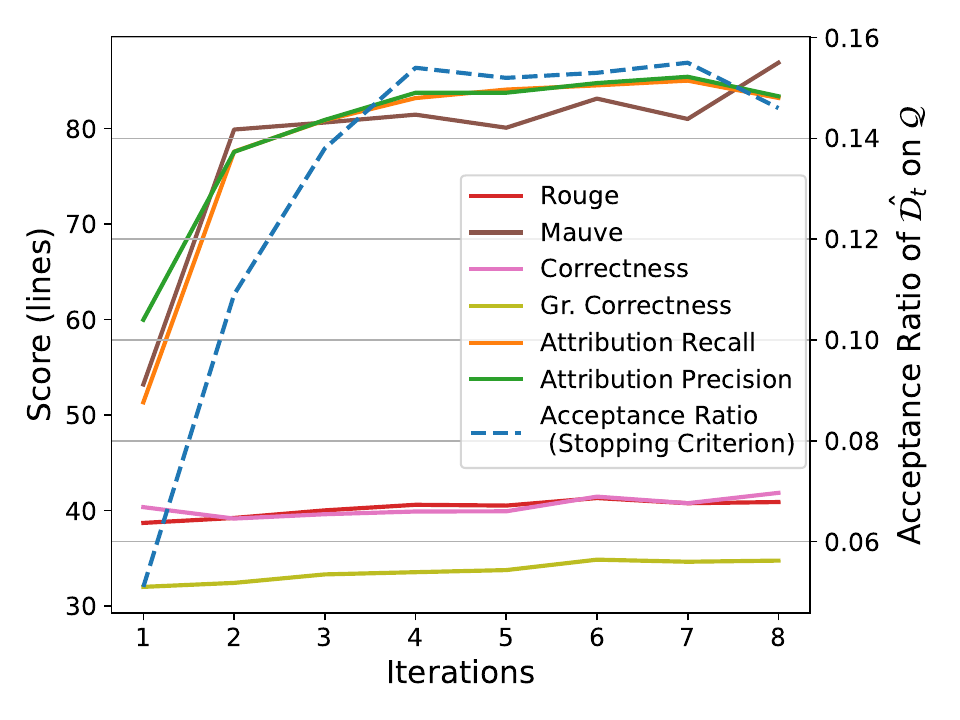}
	\caption{Evaluation metrics and \projectname's dynamic stopping criterion over the number of iterations on ASQA.}
	\label{fig:iterative-training}
\end{figure}

\paragraph{Iterative Training.}
Performance of \projectname \ across iterations is shown in Figure \ref{fig:iterative-training} on ASQA. We see a majority of citation performance improvements within the first three iterations after which citation F$_{1}$ stabilizes. %
We further observe consistent improvements to MAUVE, ROUGE-L, and grounded correctness across iterations. Importantly, we do not observe erratic or unstable performance, indicating the robustness of our iterative training procedure. The proportion of filtered examples $\mathcal{\hat{\mathcal{D}}}_k$ over $\Tilde{Y}_k$ as our dynamic stopping criterion matches the citation performance well, improving efficiency by early stopping once saturated (here iteration $4$). %

\paragraph{Selection of FCM.}
We further investigate the extent to which the quality of an FCM translates to improved citation production for \projectname. To this end, we replace the FCM %
with: (i) a DeBERTav3 model adjusted for factual consistency \citep{steen-etal-2023-little}, (ii) the current T5-based state-of-the-art on the TRUE benchmark \citep{gekhman-etal-2023-trueteacher}, (iii) the FCM \citet{honovich-etal-2022-true-evaluating} used for citation evaluation. We constrain $\hat{\mathcal{D}}_t$ to $32$ samples and run without $\mathcal{L}_{\text{FL}}$ since for decoders the computation of Shapley values over their vocabulary (prediction) is more involved. Table \ref{tab:consistency-model} shows results on the TRUE benchmark (summarization subset) and the citation F$_1$ on ASQA with \projectname. We generally observe that better scores on the TRUE benchmark translate to higher citation scores when incorporated into \projectname. The only exception is the FCM \citet{honovich-etal-2022-true-evaluating}, which is used in both training and evaluation, scoring disproportionately high in terms of citation F$_1$, most likely due to model biases leaking through evaluation. While \citet{gekhman-etal-2023-trueteacher} performs the best on TRUE, it is much more expensive to run at scale for \projectname \ than AlignScore. %

\begin{table}[ht!]
	\centering
	\resizebox{1\linewidth}{!}{
	\begin{tabular}{l | r r}
            \toprule
            Method & TRUE AUC &  Citation F$_1$\\
            \hline
            \citet{steen-etal-2023-little} & 79.5 & 72.5 \\
            \citet{honovich-etal-2022-true-evaluating}\textsuperscript{\textdagger}& 82.7 & 81.9\\
            AlignScore \citep{zha-etal-2023-alignscore} & 84.8 & 74.0  \\
            \citet{gekhman-etal-2023-trueteacher} & 87.8 &  77.7 \\
            \hline
	\end{tabular}}
	\caption{
 Measuring the extent to which the quality of an FCM (TRUE benchmark) translates to CaLF by
 comparing the quality of an FCM (TRUE benchmark) and \projectname. %
    \textsuperscript{\textdagger}Citation evaluation model.} %
        \label{tab:consistency-model}
\end{table}

\begin{table*}[ht!]
	\centering
	\resizebox{0.93\linewidth}{!}{
	\begin{tabular}{l | r r r r r r r }
            \toprule
            Method & Similarity & Fluency & Correct. & Correct. &  Citation & Citation & Avg length \\
            & Rouge-L & MAUVE & EM Rec. &  in $P$ & Recall & Precision &   \\
            \hline
            Ours  & 40.5 & 80.1 & 39.9 & 33.8 & 84.1 & 83.8 & 71.2 \\
            \hline
            Copy (No length limit) & 25.3 & 11.8 & 50.6 & 49.5 & 98.2 & 98.7 & 628.4 \\
            Copy (Truncated - 1st paragraph) & 34.0 & 16.8 & 36.0 & 34.6 & 97.9 & 98.4 & 210.7 \\
            Copy (Truncated - 100 tokens) & 33.3 & 16.2 & 22.8 & 20.9&  96.1 & 96.3 & 100.0 \\
            Ours  w/ all set to [1] & 40.5 & 80.1 & 39.9 & 33.8 & 49.0 & 49.5 & 71.2 \\
            Ours w/ cite all & 40.5 & 80.1 & 39.9 & 33.8 & 75.6 & 36.4 & 71.2 \\
            Ours w/o EOS token  & 33.0 & 20.2 & 46.9 & 39.7 & 71.1 & 81.7 & 256.0 \\
            \bottomrule
	\end{tabular}}
	\caption{Evaluation of adversarial baselines on ASQA, using MistralOrca-7B. Results are shown for seed 42. \textit{Copy}: Use retrieved passages directly as a response. %
    \textit{w/ all set to [1]}: Replacement of all citations in the generated response with citations to the first passage. \textit{w/ cite all}: Replacement of citations in the generated response with citations to all passages. \textit{w/o EOS token}: Remove EOS token from training. The average gold answer length is $113$ tokens.}
        \label{table:trivial-baselines}
\end{table*}

\paragraph{Adversarial Baselines.} Long-form QA is notoriously difficult to automatically evaluate \citep{xu-etal-2023-critical, krishna-etal-2021-hurdles}. Thus, to assess whether the automatic evaluation metrics of \citet{gao-etal-2023-enabling} can be exploited, we designed several adversarial baselines, such as copying retrieved passages as the answer or citing all passages for every generated sentence. Results are shown in Table \ref{table:trivial-baselines}. Every baseline that attempts to trick one metric performs poorly on another. For instance, copying passages (Copy) results in very poor MAUVE while citing all passages (Ours w/ cite all) results in very low citation precision. A more detailed discussion is shown in Appendix \ref{app:evaluation}.

\paragraph{Human Evaluation.}
\label{sec:human-eval}

We further conduct a human evaluation to compare the quality of generated responses produced by \projectname  \ with Few-shot FT (using MistalOrca-7B). Human subjects are tasked to judge the models' generations regarding citation quality, informativeness, coherence, and fluency, following human evaluation in \citet{gao-etal-2023-enabling} and recommendations in \citet{zhong-etal-2022-towards}. The latter three metrics are measured using a five-point Likert scale, with five being the highest score.
We randomly sample $30$ instances for each model from ELI5, ASQA, and BIO, resulting in $180$ instances. Results are shown in Table \ref{tab:human-evaluation}. \projectname \ achieves substantially higher citation quality, fluency, and answer coherence ratings than across datasets with an F$_1$ of $91.5$ versus $72.7$ for 
the baseline. While average informativeness ratings are comparable between \projectname \ and the baseline , on ELI5 particularly we observe low informativeness with \projectname. This is caused by frequent retrieval errors since ELI5 uses a weak retriever (BM25) to efficiently traverse over its large and noisy knowledge base.
Instead of using its parametric memory when passages contain little relevant information, we observe that \projectname \ still produces accurately grounded responses, however, at the cost of less informative content. %

\begin{table}[ht!]
	\centering
	\resizebox{1\linewidth}{!}{
	\begin{tabular}{l | r r r r }
            \toprule
            Method & Citation & Informa- & Coherence & Fluency \\
             &  F$_1$ & tiveness &  &  \\
            & of 100 $\uparrow$ & \multicolumn{3}{c}{Likert Scale 1 to 5 $\uparrow$} \\
            \hline
            \textbf{Few-shot }  \hspace{0.2em} ASQA & 65.2 & 4.07 & 4.07 & 4.57 \\
            \multicolumn{1}{c|}{\textbf{FT}  \hspace{3em}  ELI5} & 68.7 & 4.13 & 4.07 & 4.5 \\
            \multicolumn{1}{c|}{\hspace{4em} BIO} & 84.3 & 4.6 & 4.57 & 4.67 \\
            \hline
            \multicolumn{1}{c|}{ \hspace{4em}  Avg} & 72.7 & 4.27 & 4.23 & 4.58 \\
            \hline
            \hline
            \multicolumn{1}{c|}{\textbf{Ours} \hspace{2em}  ASQA} & 93.7 & 4.67 & 4.67 & 4.93 \\
            \multicolumn{1}{c|}{\hspace{4em} ELI5} & 82.7 & 3.3 & 3.93 & 4.17 \\
            \multicolumn{1}{c|}{\hspace{4em} BIO} & 97.3 & 4.87 & 4.77 & 4.83 \\
            \hline
            \multicolumn{1}{c|}{\hspace{4em} Avg} & \textbf{91.5} & \textbf{4.28} & \textbf{4.46} & \textbf{4.64} \\
            \hline
	\end{tabular}}
	\caption{Human evaluation study. Informativeness, coherence, and fluency are judged using a five-point Likert scale. 
 Results confirm automatic evaluation, putting \projectname \ ahead in terms of citation quality while maintaining high response quality. }
        \label{tab:human-evaluation}
\end{table}

\paragraph{Efficiency}
At inference time, \projectname \ matches the efficiency of the few-shot FT baseline, contrasting previous methods that cause significant overhead, including in-context learning (increased input length), Self-RAG (tree-decoding with critique tokens), and post-hoc editing. The training complexity of \projectname \ can be described as $\mathcal{O}(K \times |\mathcal{X})|)$, with $K$ and $\mathcal{X}$ being the number of iterations and the collection size of questions we generate weakly-supervised data of, respectively. The generation of diverse answer candidates is computationally the most involved while the filtering of answer candidates and the computation of Shapley Values is efficient due to the small size of the FCM we use (Alignscore, 355M parameters). %
To quantify the computational cost, we measured training times using a single A100 40GB GPU. We measure a training time of 13h51min for \projectname \ and 1h2min for the few-shot FT baseline. While there's a notable disparity in training time, it's essential to note that achieving comparable performance to \projectname \ via regular fine-tuning would necessitate training on significantly more data, resulting in additional training and, importantly, data annotation cost. %

\section{Conclusion}
This paper presented \projectname, a fine-tuning method for LLMs to produce accurate citations alongside generated text. It focuses on using FCMs as a training signal
by filtering candidate answers with citations and by re-weighting the LLM's objective function according to the tokens' factual importance.
\projectname \ outperforms all baselines in terms of citation quality and passage-grounded correctness while ensuring that the overall quality of responses remains high, measured via both automated and human evaluation.  In a domain transfer setting we further validate the generalizability of \projectname.
Moreover, we discuss the benefits of accurate citation production for improving factuality and highlight the importance of each component of \projectname \ through a systematic ablation. Future work looks at %
incorporating a learned mechanism into \projectname \ to abstain from answering if none of the retrieved passages are considered relevant to the question.

\newpage
\section*{Limitations}
The assumption that every generated sentence requires a citation is an oversimplification we adopt from \citet{gao-etal-2023-enabling}. However, real-world dialogue agents commonly introduce their response with non-factual phrases or sentences (e.g. \emph{Of course I can help!}). A potential solution is to introduce an extrapolatory label as described in \citet{yue-etal-2023-automatic}. Furthermore, \projectname \ is not entirely model-agnostic out-of-the-box, since the tokenization alignment algorithm was designed with the idiosyncrasies of the particular LLMs and FCMs in mind. Moreover, inaccuracies or biases of the factual consistency models could negatively affect the citation quality of \projectname \ (c.f.\ Sec.~\ref{sec:discussion}). A common strategy to mitigate biases involves an alignment stage, where models are explicitly optimized on based on preference data. It would be interesting to explore alignment in the context of \projectname \ applied to highly domain-specific questions, such as those compiled in the recent dataset of \citet{malaviya2024expertqa}.

Finally, our paper focuses exclusively on improving generation, yet as highlighted in Section \ref{sec:human-eval}, high-quality retrieval systems are vital for citations to be effective. When retrieved passages contain non-factual information or lack useful content, produced answers might not be informative. While the reliance of retrieved passage is answer production poses a limitation, language model's intrinsic memory for content generation forfeits the transparency and verifiability benefits inherent in citation production. Balancing the use of citation production and intrinsic memory while preserving the advantages of both remains an ongoing challenge. Future directions might consider a joint fine-tuning procedure of generator and retriever and even capturing their rich interactions, such as \emph{abstaining} from answering questions when retrieved passages are irrelevant. The action of abstaining to answer could serve as a signal to the retrieval system to seek alternative sources of information.

\section*{Ethics Statement}

Our paper improves the ability of LLMs to produce accurate citations to sources alongside their generated answers to improve the verifiability of LLMs' responses. As noted in Section \ref{sec:factuality}, we caution against equating improved citation quality with improved factuality since even correctly cited passages can be factually incorrect or misleading without additional context about the passage and its source. Citations can therefore invoke a false sense of trust and amplify observations made in \citet{si2023large}. It remains important for users to stay critical and reflect on generated responses. By improving citation accuracy our paper simplifies the verification process but does not eliminate it. Furthermore, \citet{huang2024citation} argue that the incorporation of citations can help to address intellectual property and associated ethical issues in the deployment of LLMs due to the increased verifiability of responses made. Finally, we recognize a potential risk of dual-use by adversarial actors: similarly to how humans cherry-pick data to support a strong bias or prior, we cannot guarantee that an LLM will not exhibit similar behaviour when manipulated with passages that contain misinformation or miss relevant context.

\section*{Acknowledgements}
The authors would like to thank the anonymous reviewers for their time and detailed feedback on our paper. We further thank Tony Hu for being part of the human evaluation.

\bibliography{anthology, custom}

\clearpage

\newpage

\appendix

\section{Appendix}
\label{sec:appendix}

\subsection{Datasets} 
\label{app:datasets}
The four datasets considered for evaluation cover different question-answering tasks and domains: ASQA is a disambiguation task built on top of AmbigQA \citep{min-etal-2020-ambigqa}, ELI5 contains real-world highly open-ended questions and answers from an online forum (Reddit), Hagrid contains entity-specific questions, and BIO is a biography generation task containing simple person-related questions. Hagrid's training set as provided by \citet{kamalloo2023hagrid} consists of 1,922 GPT-4 generated answers, out of which 335 instances are human-labeled as both attributable and informative. Only these $335$ instances are used as training data.  For Hagrid, we randomly sample $4$ instances for \projectname \ and use all its $335$ training samples for our fine-tuning baseline. We use each datasets' original training set to sample questions $\mathcal{X}$ and atomic facts $A$. Since Hagrid does not have any annotations for factual coverage, we compute the quality assurance function $\mathcal{Q}$ only over the Citation conditions. BIO consists of two evaluation sets. As recommended by the authors and following Self-RAG, we use the second, and more difficult, evaluation set in our experiments. The test set of ALCE-ASQA and ALCE-ELI5 consists of $1000$ randomly sampled instances from their respective development set. For more details on the ALCE benchmark, we refer to \citet{gao-etal-2023-enabling}.

\subsection{Experimental Setup}
\label{app:experimental-setup}

\paragraph{Token alignment algorithm} We design the token alignment algorithm around the idiosyncrasies of the LLMs' and FCM's tokenizers. Llama2, Mistral-Instruct, and MistralOrca use the Llama tokenizer\footnote{\url{https://huggingface.co/docs/transformers/main/en/model_doc/llama\#transformers.LlamaTokenizer}}, a BPE model \citep{sennrich-etal-2016-neural} based on sentencepiece\footnote{\url{https://github.com/google/sentencepiece?tab=readme-ov-file}}. Similarly, the FCM, Alignscore, uses a RoBERTa tokenizer which is derived from the GPT-2 tokenizer, also using BPE. Both Llama and RoBERTa tokenizers treat spaces as parts of the tokens. In contrast to RoBERTa tokenizer, the Llama tokenizer indicates each space via an underscore ("\_"). We exclude special tokens of either tokenizer (e.g. "<s>" or "<0x0A>") from the alignment procedure. 

Given a sentence, tokenized both by FCM and LLM, we %
we first strip the FCM tokens (i.e. remove spaces) and remove underscores for tokens of the LLM, essentially removing all spaces to unify their representations. The algorithm then checks whether the current tokens are equal or subsets of one another, with and without considering the memory of tokens not yet aligned but iterated over by the pointers.  For instance, consider the FCM tokens: 'Maw', 'syn', 'ram', and LLM tokens for the same word: is \_M', 'aws', 'yn', 'ram'. For the first two FCM tokens, none of the LLM tokens can be directly aligned to it. Our algorithm ensures to find the \emph{smallest sequence} of tokens in both the LLM and FCM that can be aligned to each other (here the spans are 'Mawsyn' and 'ram') and assigns the relevance score as described in Section \ref{sec:focused-learning}. We find the smallest sequence by keeping a list of tokens for each FCM and LLM that could not yet be aligned and increment the respective pointers according to which sequence needs to continue (e.g. Checking that [M] + [was] continues [Maw] potentially, requires incrementing the FCM pointer, since an 's' is missing in [Maw]. This algorithm is efficient and runs in linear time.

\paragraph{Generating Cited Answers for Training.} While datasets such as ELI5 have over 250K training instances we could use to generate weakly-supervised training data, we constrained the size of $\mathcal{X}$ to $1000$ samples, since it is computationally infeasible to run our weakly-supervised data generation procedure on all training samples of the dataset. The threshold $\Theta$ is determined dynamically. Starting with a value of $0.9$, if the number of samples in $\hat{D}$ is below $3$, we consider the threshold too high for the given LLM and reduce it by $0.1$ until the requirement is met.

\paragraph{Passage Retrieval.} For the retriever $\mathcal{R}$ we use GTR \citep{ni2021large} for ASQA and Hagrid (Wikipedia), BM25 for ELI5 (CommonCrawl\footnote{http://commoncrawl.org}), and Contriever-MS MARCO \citep{izacard2022unsupervised} for BIO (Wikipedia), to maintain comparability with \citet{gao-etal-2023-enabling} and \citet{asai2023self}. We use $|P|=3$ passages throughout due to context length limitations. We use the same indices as provided by ALCE, subsequently, each passage has a length of $100$ tokens.

\paragraph{Answer Generation.} We use the same instructions/prompts across all models, specifically the \texttt{default} prompts from ALCE. In contrast to the code of ALCE\footnote{https://github.com/princeton-nlp/ALCE}, we used the chat templates across all models and baselines\footnote{\url{https://huggingface.co/docs/transformers/main/en/chat_templating}} to render the inputs, since we observed crucial tokens missing during the fine-tuning procedure otherwise. The only exception is ELI5, where we observed that chat templates resulted in worse performance and used ALCE's strategy plus relevant formatting tokens instead across all models and baselines. For inference, we use MAP with a beam size of $1$, instead of sampling with temperature scaling as done in ALCE, assuming that this results in more factual outputs.\footnote{Note that we do not use constrained decoding for citation generation despite its apparent suitability here. Yet, the position of citation markers within a sentence can be rather flexible, which we consider a desired property for naturally produced text.}

\paragraph{Factual Consistency Model} Our choice of using AlignScore as our FCM $\phi$ was further motivated by its dissimilarity to the citation evaluation model \citep{honovich-etal-2022-true-evaluating} (in contrast to e.g. TrueTeacher \citep{gekhman-etal-2023-trueteacher} which in principle performed better but is more similar to the evaluation model). While AlignScore is an encoder-based RoBERTa model, TRUE \citep{honovich-etal-2022-true-evaluating} is an encoder-decoder T5-11B model. Moreover, their training data is different. AlignScore was trained on 15 datasets of various tasks, including natural language inference, summarization, information retrieval, and paraphrasing while TRUE has only seen $6$ natural language inference-related datasets.

\paragraph{Training \& Hyperparameters} We set the learning rate to $3^{-4}$ and train for a total of $100$ steps across all models and experiments. The maximum generation length is set to $256$ tokens, however, most generations stay far below this limit (see Appendix \ref{app:evaluation}. We use adamw \citep{loshchilov2018decoupled} as the optimizer. We use a batch size of $1$ during training with gradient accumulation, resulting in an effective batch size of $4$. For LoRA, we use a rank $r=4$ and apply it to all parts of the attention mechanism. For fine-tuning, we exclude tokens of the prompts from the loss computation that are not part of the gold answer, so we are not fine-tuning the instructions, only the answers that follow after the instruction. %
The iterative training process stops after at most $8$ iterations due to computational constraints. We further down-weight the loss contribution of the EOS token to $0.02$ since models tend to otherwise produce very short single-sentence responses.

\paragraph{Implementation Details} We use the Huggingface checkpoints for LLama2-7B\footnote{\url{https://huggingface.co/meta-llama/Llama-2-7b-chat-hf}}, MistralOrca-7B\footnote{\url{https://huggingface.co/Open-Orca/Mistral-7B-OpenOrca}}, and Mistral-Instruct-7B,\footnote{\url{https://huggingface.co/mistralai/Mistral-7B-Instruct-v0.1}} as well as for all FCMs we considered in our experiments, namely AlignScore (RoBERTa-large, 355M parameters) \footnote{\url{https://huggingface.co/yzha/AlignScore}, and their repository \url{https://github.com/yuh-zha/AlignScore}}, TRUE (T5, 11B parameters)\footnote{\url{https://huggingface.co/google/t5_xxl_true_nli_mixture}}, \citep{steen-etal-2023-little}\footnote{\url{https://huggingface.co/juliussteen/DeBERTa-v3-FaithAug}} (DeBERTaV3, 304M parameters), and TrueTeacher (T5, 11B parameters)\footnote{\url{https://huggingface.co/google/t5_11b_trueteacher_and_anli}}. The Mistral models are licensed under Apache2.0, AlignScore and \citep{steen-etal-2023-little} are licensed under MIT, TrueTeacher is licensed under cc-by-nc-4.0, and Llama2 is licensed under the llama license\footnote{\url{https://github.com/facebookresearch/llama/blob/main/LICENSE}}. Subsequently, our research is consistent with the licenses' intended use. The models are intended to be used for English. We use the ALCE data and prompts from their repository\footnote{\url{https://github.com/princeton-nlp/ALCE}}, and Huggingface's Datasets for Hagrid\footnote{\url{https://huggingface.co/datasets/miracl/hagrid}}. For running the experiments, we used a combination of A100 40GB and A10G with 23GB GPUs. We use the Python package rouge-score\footnote{\url{https://pypi.org/project/rouge-score/}} for computing the ROUGE, and the package mauve-text\footnote{\url{https://pypi.org/project/mauve-text/}} for computing the MAUVE score. We adopt the code of ALCE for the citation and correctness metrics and define passage-grounded correctness ourselves. We use NLTK \citep{bird-loper-2004-nltk} for some pre-and post-processing steps.

\paragraph{Baselines.} Results from state-of-the-art methods are taken from their respective papers. The only exception is Self-RAG on ELI5, which we have run by ourselves using the authors' repository and their models\footnote{https://github.com/AkariAsai/self-rag}. At the point of submission, their repository produces results which are worse than reported in their paper, as explained by the authors due to a bug\footnote{https://github.com/AkariAsai/self-rag/issues/4}, which potentially impacts scores reported of Self-RAG on ELI5. While the authors have stated their intentions to fix this issue, the problem was not resolved as of our submission date. We will update their scores once a fix is available.

\subsection{Evaluation}
\label{app:evaluation}

Detailed descriptions for each evaluation metric are shown in Table \ref{tab:metrics}. Results shown in Table \ref{tab:domain-transfer} are computed with default random seed $42$.

\begin{table*}[ht!]
	\centering
	\resizebox{1\linewidth}{!}{
	\begin{tabular}{l | p{7cm} | p{10cm}}
            \toprule
            Name & Description & Computation \\
            \hline
            \multicolumn{3}{c}{\cellcolor{gray!25} Gold Answer-based Metrics} \\
             \hline
             \textbf{Fluency} (MAUVE) & Measures two types of errors: (i) model produces degenerate text (outside of human distribution), (ii) model does not yield diverse text (does not cover human distribution)  & $R_{\lambda} = \lambda P + (1-\lambda) Q$, summarizing KL divergence of $\text{KL}(P | R_{\lambda}$) and $\text{KL}(Q | R_{\lambda}$, for $\lambda \in (0,1)$, computed via Monte-Carlo estimator (LM embeddings + quantization via k-means).\\ 
             \hline
             \textbf{Similarity} (ROUGE-L)  & Measures longest matching sequence of words. Bad approximator for factuality (see e.g. \href{https://arxiv.org/pdf/2305.18201.pdf}{here}) & Computes $F_{1}$ for longest common subsequence (LCS). Recall being ratio over reference. Precision ratio over answer. \\
             \midrule
             \multicolumn{3}{c}{\cellcolor{gray!25} Specialized Metrics} \\
             \midrule
            \textbf{Correctness} & Measures whether atomic units of information from gold answers appear in the generated answer. & ASQA: Measures exact match of short answers in generated answers. ELI5: Measures whether atomic statements are consistent with generated answer. \\
            \hline
            \textbf{Passage-grounded Correctness} & Measures whether atomic units of information from gold answers appear in the generated answers \emph{and} whether this information is supported by the retrieved passages. Eliminates to score responses that are correct but not grounded in retrieved passages. & ASQA: Measures exact match of short answers in generated answers that can be attributed to retrieved passages. ELI5: Measures whether atomic statements are consistent with generated answer that can be attributed to retrieved passages . \\
            \hline
            \textbf{Citation Recall} & Measures the ratio of sentences in generated answers that are consistent with their cited sources/passages.  & Recall is 1 iff there exists at least one citation and sentence is consistent with citations considered jointly: $\phi_{eval}(\oplus(C_{i}), s_{i})=1$.\\
            \hline
            \textbf{Citation Precision} & Measures the ratio of citations in generated answer that are not irrelevant. & Citation considered irrelevant iff: i) the citation itself does not support the attributed sentence:  $\phi_{eval}(c_{i,j}, s_{i})=0$ ii) removing the citation does not affect rest of citations to attribute the sentence:  $\phi_{eval}(\oplus(C_{i}) \text{\textbackslash} \{c_{i,j}\}, s_{i})=1$. \\
            \bottomrule
	\end{tabular}}
	\caption{Overview of the LFQA evaluation metrics. Correctness, Fluency, and Citation scores are taken from the ALCE \citep{gao-etal-2023-enabling}. Passage-grounded Correctness is a metric we propose to measure information coverage for citable statements (i.e. excluding hallucinated correct information). $\phi_{\text{eval}}$ is the evaluation FCM, namely TRUE \citep{honovich-etal-2022-true-evaluating}.}
    \label{tab:metrics}
\end{table*}

\paragraph{Ablation}

Table \ref{tab:ablation-mistral-instruct} Table \ref{tab:ablation-llama2}  show the ablation results for \projectname on Mistral-Instruct 7B and Llamav2-7B-chat, respectively. The results largely align with observations made for MistralOrca-7B in table \ref{tab:ablation}. We observe that passage-grounded correctness either slightly decreases or remains comparable to the baseline when using the weakly-supervised training without our focused learning objective $\mathcal{L}_{\text{FL}}$. Once the objective is added,vcorrectness improves consistently across datasets and models over the baseline.

\begin{table}
	\centering
	\resizebox{1\linewidth}{!}{
	\begin{tabular}{l | r r r}
            \toprule
             & Correctness &  Citation  & Citation \\
            Method & in $P$ & Recall & Precision \\
            \hline
            \multicolumn{4}{c}{\cellcolor{gray!25} ASQA } \\
            LLM & 28.3 & 60.9 & 63.5 \\
            LLM + WS & 24.1 (\textcolor{red}{-4.2}) & 72.3 (\textcolor{applegreen}{+11.4}) & 73.9 (\textcolor{applegreen}{+10.4})\\
            LLM + WS + $\mathcal{L}_{FL}$ & \textbf{29.6} (\textcolor{applegreen}{+1.3}) & \textbf{79.2} (\textcolor{applegreen}{+19.7}) & \textbf{80.2} (\textcolor{applegreen}{+16.7})\\
            \multicolumn{4}{c}{\cellcolor{gray!25} ELI5 } \\
            LLM & 9.1 & 55.2 & 40.1 \\
            LLM + WS & 10.9 (\textcolor{applegreen}{+1.8}) & 60.9 (\textcolor{applegreen}{+5.4}) & 59.6 (\textcolor{applegreen}{+19.5}) \\
            LLM + WS + $\mathcal{L}_{FL}$ & \textbf{12.5} (\textcolor{applegreen}{+3.4}) & \textbf{70.0} (\textcolor{applegreen}{+14.8}) & \textbf{67.0} (\textcolor{applegreen}{+26.9}) \\    
            \hline
	\end{tabular}}
	\caption{Ablation for \projectname \ (Mistral-Instruct-7B). WS: weakly-supervised training, $\mathcal{L}_{FL}$: Focused learning.}
        \label{tab:ablation-mistral-instruct}
\end{table}

\begin{table}
	\centering
	\resizebox{1\linewidth}{!}{
	\begin{tabular}{l | r r r}
            \toprule
             & Correctness &  Citation  & Citation \\
            Method & in $P$ & Recall & Precision \\
            \hline
            \multicolumn{4}{c}{\cellcolor{gray!25} ASQA } \\
            LLM & 23.4 & 58.7 & 55.3 \\
            LLM + WS & 23.1 (\textcolor{red}{-0.3}) & 69.6 (\textcolor{applegreen}{+10.9}) & 66.3 (\textcolor{applegreen}{+11.0})\\
            LLM + WS + $\mathcal{L}_{FL}$ & \textbf{30.7} (\textcolor{applegreen}{+7.3}) & \textbf{76.0} (\textcolor{applegreen}{+17.3}) & \textbf{72.5} (\textcolor{applegreen}{+17.2})\\
            \multicolumn{4}{c}{\cellcolor{gray!25} ELI5 } \\
            LLM & 11.3 & 53.2 & 46.6 \\
            LLM + WS & 9.1 (\textcolor{red}{-2.1}) & 67.1 (\textcolor{applegreen}{+13.9}) & 66.4 (\textcolor{applegreen}{+19.8}) \\
            LLM + WS + $\mathcal{L}_{FL}$ & \textbf{11.7} (\textcolor{applegreen}{+0.4}) & \textbf{71.2} (\textcolor{applegreen}{+18.0}) & \textbf{63.2} (\textcolor{applegreen}{+16.6}) \\    
            \hline
	\end{tabular}}
	\caption{Ablation for \projectname \ (Llama2-7B-chat). WS: weakly-supervised training, $\mathcal{L}_{FL}$: Focused learning.}
        \label{tab:ablation-llama2}
\end{table}

\begin{table*}[ht!]
	\centering
	\resizebox{1\linewidth}{!}{
	\begin{tabular}{l | r r r r r r r r }
            \toprule
             Method & Rouge-L & Fluency & Correctness & Grounded Correct. & Citation Recall & Citaion Precision & Avg length \\
            \hline
            \multicolumn{8}{c}{\cellcolor{gray!25} ASQA } \\
            ChatGPT \citep{gao-etal-2023-enabling} & -- & 	66.6 & \textbf{40.4} & -- &  	73.6&	72.5 & --\\
            Vicuna-13B \citep{gao-etal-2023-enabling} & -- & 82.6 & 31.9 & -- & 51.1 & 50.1 & -- \\
            Self-RAG 7B \citep{asai2023self} & 35.7&	74.3 & 30.0 & -- &	66.9&	67.8 & --\\
            \hline
            In-context (Llamav2-7B-chat) & 35.9  & 84.1 & 34.5 & 25.2 & 50.4 & 50.0 & 93.3 \\
            Few-Shot FT (Llamav2-7B-chat) & 34.7 & 71.3 & 33.1 & 23.4 & 58.7 &  55.3 & 52.5 \\
            In-context (Mistral-Instruct 7B) & 36.4 & 86.6 & 34.1 & 28.0 & 21.7 & 23.6 & 75.9 \\
            Few-Shot FT (Mistral-Instruct 7B) &  37.6 & 82.5 & 35.3 & 28.3 & 60.9 & 63.5 & 105.7 \\
            In-context (MistralOrca 7B) &  38.8 & 53.9 & 40.1 & 32.4 & 52.5 & 61.0 & 64.5\\
            Few-Shot FT (MistralOrca 7B) & 37.5 & 81.9 & 37.3 & 25.8 & 57.5 & 55.2 &  61.0\\
            \hline
            Ours (Llamav2-7B-chat) & 37.9 & 84.3 & 37.8 & 29.5 & 72.8 & 72.3 & 86.9 \\
            Ours (Mistral-Instruct 7B) & 37.7 & \textbf{87.5} & 35.2 & 29.6 & 79.2 & 80.2 & 79.5 \\
            Ours (MistralOrca 7B) & \textbf{40.5} & 80.1 & 39.9 & \textbf{33.8} & \textbf{84.1} & \textbf{83.8} & 71.2 \\
            \multicolumn{8}{c}{\cellcolor{gray!25}  ELI5 } \\
             ChatGPT \citep{gao-etal-2023-enabling} & -- & 	57.2 & 12.0 & -- &  51.1 &	50.0 & --\\
              Vicuna-13B \citep{gao-etal-2023-enabling} & -- & 58.2 & 10.0 & -- & 15.6 & 19.6 & -- \\
             Self-RAG 7B \citep{asai2023self} & 16.9 & 32.6 & 9.7 & 5.4 & 23.3 & 33.9 & --\\
            \hline
            In-context (Llamav2-7B-chat)  & 19.7 & 37.7 & 14.1 & 8.6  &  39.9 & 27.6 & 110.7 \\
            Few-Shot FT (Llamav2-7B-chat) & 21.3 & 54.9 & 17.8 & 11.3 & 53.2 & 46.6  & 138.4 \\
            In-context (Mistral-Instruct 7B) & 20.5 & 62.3 & 17.4 & 11.7 & 43.8 & 44.9 & 111.2 \\
            Few-Shot FT (Mistral-Instruct 7B) & 19.5 & 37.8 & 13.8 & 9.1 & 55.2 & 40.1 & 93.0 \\
            In-context (MistralOrca 7B)  & 20.8 & 27.7 & \textbf{20.5} & 12.4 & 45.4 & 41.8 & 94.8 \\
            Few-Shot FT (MistralOrca 7B)  & 20.5 & 44.9 & 18.4 & 11.0 & 57.3 & 51.0 & 100.8 \\
            \hline
            Ours (Llamav2-7B-chat)  & 21.3 & \textbf{69.5} & 17.2 & 11.7 & 71.2 & 63.2 &  141.2 \\
            Ours (Mistral-Instruct 7B) & \textbf{21.8} & 53.5 & 18.9 & \textbf{12.5} & 70.0 & \textbf{67.0} & 143.8 \\
            Ours (MistralOrca 7B)  & 20.7 & 68.3 & 18.6 & \textbf{12.5} & \textbf{72.1} & 66.6 & 108.3 \\
            \bottomrule
	\end{tabular}}
	\caption{Main in-domain results on ASQA, ELI5 using \projectname \ for fine-tuning various instruction-tuned LLMs, using only $4$ initial samples $\mathcal{D}$. Results are shown for default random seed $42$.}
    \label{tab:main-results-app}
\end{table*}

\begin{table*}[ht!]
	\centering
	\resizebox{1\linewidth}{!}{
	\begin{tabular}{l l | r r r r r r r }
            \toprule
            Method & Source &  Rouge & Fluency & Correct. & Gr. Correct. & Attr. Recall & Attr. Precision & Length \\
            \hline
            \multicolumn{9}{c}{\cellcolor{gray!25} Target Dataset: ASQA } \\
            Self-RAG 7B & Critique tokens & 35.7 & 74.3 & 30.0 & -- &	66.9&	67.8 & --\\
            \hline
            Zero-Shot (Llama2-Chat-7B) & -- & 36.1 & 47.5 & 35.6 & 27.1 & 24.3 & 42.8 & 126.9 \\
            Zero-Shot (MistralOrca) & -- & 39.0 & 78.9 & 39.5 & 31.6 & 5.3 & 6.1 & 63.6\\
             Few-shot FT (Llama2-Chat-7B) & ELI5 & 37.2 & 74.0 & 37.7 & 31.6 & 67.9 & 62.2 & 150.6 \\
             Few-shot FT (MistralOrca) & ELI5  & 39.7 & \textbf{90.1} & 38.5 & 31.4 & 73.8 & 69.7 & 94.1 \\
             FT (MistralOrca) & Hagrid & 36.7 & 66.1 & 37.8 & 29.5 & 50.8 & 51.8 & 57.6\\
            \hline
            Ours (Llama2-Chat-7B) & ELI5 &37.1 & 75.7 & 36.1 & 30.4 & 77.8 & 69.0 & 141.6 \\
            Ours (MistralOrca) & ELI5 & \textbf{40.1} & 86.6 & \textbf{40.0} & \textbf{33.2} & \textbf{80.4} & 78.5 & 101.0 \\
            Ours (MistralOrca) & Hagrid & 39.7 & 80.8 & 38.6 & 32.3 & 78.9 & \textbf{81.1} & 64.9 \\
            \multicolumn{9}{c}{\cellcolor{gray!25} Target Dataset: ELI5} \\ 
            Self-RAG 7B & Instr. tuning w/ critique tokens & 16.9 & 32.6 & 9.7 & 5.4 & 23.3 & 33.9 & --\\
            \hline
            Zero-Shot (Llama2-Chat-7B) & --  & 20.0 & 26.5 & 15.3 & 9.7 & 17.8 & 39.0 & 111.1\\
            Zero-Shot (MistralOrca) & -- & \textbf{21.3} & 35.0 & \textbf{22.2} & 12.6 & 8.1 & 14.4 & 120.5\\
            Few-shot FT (Llama2-Chat-7B) & ASQA & 17.1 & 20.8 & 11.4 & 6.8 & 33.3 & 32.6 & 70.4 \\
            Few-shot FT (MistralOrca)  & ASQA & 20.9 & 41.1 & 19.7 & 10.6 & 39.8 & 41.0 & 92.1 \\
            FT (MistralOrca) & Hagrid & \textbf{21.3} & \textbf{58.1} & 20.9 & 12.7 & 28.6 & 37.0 & 112.4 \\
            \hline
            Ours (Llama2-Chat-7B) & ASQA & \textbf{21.3} & 35.0 & 18.2 & 11.0 & 38.6 & 34.7 & 137.7 \\
            Ours (MistralOrca) & ASQA & 21.2 & 31.3 & 20.4 & 12.5 & \textbf{57.4} & \textbf{57.2} & 97.2 \\
            Ours (MistralOrca) & Hagrid  & 21.1 & 32.3 & 20.2 & \textbf{12.9} & 54.0 & 55.3 & 101.7\\
            \bottomrule
	\end{tabular}}
	\caption{Results for our zero-shot domain transfer setting, when trained on a source dataset and evaluated on a different target dataset without any in-context instances. Results are shown for default random seed $42$.}
    \label{tab:domain-transfer-app}
\end{table*}

\paragraph{Adversarial Baselines}
The automated metrics proposed by \citet{gao-etal-2023-enabling} do not explicitly control for the generation of irrelevant information (i.e. factual precision, such as FactScore)\footnote{The exclusion of such metric in ALCE is likely due to gold answers not being designed to be factually comprehensive.}. Subsequently, their metrics favour longer responses for coverage-based measures (i.e. correctness), as also pointed out by \citet{asai2023self}. This raises the question of whether the automated metrics deployed can be tricked with trivial responses or certain response patterns. 

Table \ref{table:trivial-baselines} shows results across metrics for several such baselines on ASQA. Our approach has an average response length of $71.2$ tokens, comparable to typical fine-tuning with $61.0$ tokens, both being substantially shorter than the dataset's average gold answer length with $113$ tokens. In contrast, the retrieved passages themselves are $521.8$ tokens long. Considering these as the answer themselves, we indeed achieve a higher correctness score ($50.6$) while maintaining perfect citation, however, we observe a substantial decrease in both ROUGE-L and MAUVE scores. This is intuitive since the retrieved passages are very dissimilar in style from the gold answers. Moreover, MAUVE has an explicit length bias \citep{pillutla2021mauve}. When explicitly biasing our \projectname \ to generate long sequences by removing the EOS token during fine-tuning and setting the generation limit to 256 tokens, we also observe a substantial increase in correctness while maintaining much of the attribution performance. Yet, again we observe a substantial decrease in ROUGE-L and MAUVE, highlighting the importance of maintaining all performance metrics high while optimizing attribution, as achieved by \projectname.
Finally, we consider replacing the citations made in generated responses from our model with: (i) citations to exclusively the first passage, (ii) citations to all passages. As seen in the table, citation scores are substantially worse than our approach. While in principle (ii) should present an upper bound on attribution recall, we observe lower scores than our model even here. This can be explained by inaccuracies in the evaluation model, being biased towards information at the beginning of the premise (scoring much worse for citations to the last passage). %

\subsection{Human Evaluation}
\label{app:human-evaluation}

Human subjects are tasked to judge the models' generations regarding: (i) citation recall: judgement whether a generated sentence is fully supported by citations, (ii) citation precision: whether a citation partially or fully supports a sentence, (iii) informativeness: whether the generation helps to answer the question, (iv) coherence: whether generated sentences are semantically and syntactically well-connected, (v) fluency: whether all sentences are grammatically correct and well-readable. While (i), (ii), and (iii) are adopted from the human evaluation in \citet{gao-etal-2023-enabling}, we further measure (iv) and (v) as recommended in \citet{zhong-etal-2022-towards}. We randomly sample $30$ instances for each model for each dataset (ELI5, ASQA, BIO), resulting in $180$ instances which are judged via the above criteria. The instances were judged by four subjects, distributed equally (45 samples per subject). Each subject was provided with detailed annotation guidelines, providing examples for each possible annotation option with explanations. The subjects were partially authors and partially volunteers. All subjects were informed how the data would be used and provided consent. All subjects are male under $65$ years of age and either from the USA or the UK.

\subsection{Qualitative Examples}

To qualitatively compare the produced answers by \projectname \ and our few-shot FT baseline we show a randomly selected output for each evaluation dataset. Results are shown in Figure \ref{app:asqa-example}, Figure \ref{app:eli5-example}, and Figure \ref{app:bio-example} for ASQA, ELI5, and BIO, respectively. 

 \begin{figure*}[ht!]
            \centering
            \fbox{\begin{minipage}{40em}
            \textbf{Question}: What is the genre of holes by louis sachar?
            
            \textbf{Retrieved Passages}: 

            [1] Title: Louis Sachar
            Text: Louis Sachar Louis Sachar ( ; born March 20, 1954) is an American young-adult mystery-comedy author. He is best known for the "Wayside School" series and the award-winner, "Holes". "Holes" won the 1998 U.S. National Book Award for Young People's Literature and the 1999 Newbery Medal for the year's "most distinguished contribution to American literature for children". In 2013, it was ranked sixth among all children's novels in a survey published by "School Library Journal". After graduating from high school, Sachar attended Antioch College for a semester before transferring to University of California, Berkeley, during which time he began helping \\
            
            [2] Title: Holes (novel)
            Text: American teenager with a criminal history. A new friendship with Ginny, who has cerebral palsy, a reunion with former friend X-Ray, a ticket-scalping scheme, a beautiful pop singer, and a frame-up all test Armpit’s resolve to "Just take small steps and keep moving forward". Holes (novel) Holes is a 1998 young adult mystery comedy novel written by Louis Sachar and first published by Farrar, Straus and Giroux. The book centers around an unlucky teenage boy named Stanley Yelnats, who is sent to Camp Green Lake, a juvenile corrections facility in a desert in Texas, after being falsely accused of theft. \\
            
            [3] Title: Holes (novel)
            Text: Holes (novel) Holes is a 1998 young adult mystery comedy novel written by Louis Sachar and first published by Farrar, Straus and Giroux. The book centers around an unlucky teenage boy named Stanley Yelnats, who is sent to Camp Green Lake, a juvenile corrections facility in a desert in Texas, after being falsely accused of theft. The plot explores the history of the area and how the actions of several characters in the past have affected Stanley's life in the present. These interconnecting stories touch on themes such as racism, homelessness, illiteracy, and arranged marriage. The book was both a\\

            \noindent\rule{40em}{0.4pt}

            \textbf{Few-shot FT Baseline Response}: "Holes by Louis Sachar is a genre of young adult mystery comedy novel [2]. It was first published by Farrar, Straus and Giroux and won the 1998 U.S. National Book Award for Young People's Literature and the 1999 Newbery Medal for the year's \"most distinguished contribution to American literature for children\" [1]. The novel centers around an unlucky teenage boy named Stanley Yelnats, who is sent to Camp Green Lake, a juvenile corrections facility in a desert in Texas, after being falsely accused of theft [2]."\\

            \textbf{\projectname \ Response}: The genre of "Holes" by Louis Sachar is a young adult mystery-comedy [1]. The novel explores themes such as racism, homelessness, illiteracy, and arranged marriage [3].

            \end{minipage}}
        \caption{ASQA Example of generated answers with citations given a question and retrieved passages. We compare the answers produced by the Few-shot FT baseline model with those generated by \projectname.}
        \label{app:asqa-example}
\end{figure*}

 \begin{figure*}[ht!]
            \centering
            \fbox{\begin{minipage}{40em}
            \textbf{Question}: Why are soap operas instantly identifiable.
            
            \textbf{Retrieved Passages}: 
            
            [1] Title: New Soap Chronicles of the Sun To Debut On TV5Monde USA - Soap Opera Spy Text: with Soap Opera Spy for all the latest news, updates and spoilers on all of your favorite daytime and primetime television shows right here! Young and the Restless Sharon Case Talks Mariah-Sharon Drama and Why Nick Is The One Days Of Our Lives Spoilers: JJ Pays It Forward – Talks New Nurse Out Of Suicide, Instantly Smitten With Her\\

            [2] Title: TV Soap Opera - English, Belonging Text: Number 96, which set a new standard and wholly different tone for Australian TV 'soapies’. Other popular and successful local soap operas included The Box (1974), The Young Doctors (1976), The Restless Years (1977), and Prisoner (1979). Teacher to survey the class and lead class discussion: What television soap operas do you watch and why? What soap character do you most closely identify with and why? Viewing questions: Megan Spencer describes the new phase of Neighbours as breaking with the old tradition of “marketing ourselves back to ourselves”. How important do you think it is for viewers to see their\\

            [3] Title: Framing The Hobbit – The Golden Lasso Text: per second. What does this mean? Well normal movies and TV shows are shown at 24 frames per second and our minds naturally fill in the missing bits. We don’t even realize that it’s happening. However, soap operas, home movies, and other videos are shot at 30 frames per second. This is why you can instantly tell a soap opera or reality TV show from other programs just by looking at it for a few seconds. In general, our minds link video and high frame rates to low budget programs. Despite Peter Jackson’s assurance that it will revolutionize movies, many\\

            \noindent\rule{40em}{0.4pt}

            \textbf{Few-shot FT Baseline Response}: Soap operas are instantly identifiable because of their distinct tone and style, which is often characterized by a specific storyline and target audience [1]. Additionally, soap operas are easily distinguishable from other programs by their frame rate, which is 30 frames per second, as opposed to the standard 24 frames per second [3].\\

            \textbf{\projectname \ Response}: Soap operas are instantly identifiable because they are shown at 30 frames per second, which is different from the 24 frames per second used for normal movies and TV shows [3]. Our minds naturally fill in the missing bits, making it easy to tell a soap opera or reality TV show from other programs just by looking at it for a few seconds [3].

            \end{minipage}}
        \caption{ELI5 Example of generated answers with citations given a question and retrieved passages. We compare the answers produced by the Few-shot FT baseline model with those generated by \projectname.}
        \label{app:eli5-example}
\end{figure*}

 \begin{figure*}[ht!]
            \centering
            \fbox{\begin{minipage}{40em}
            \textbf{Question}: Tell me a bio of Riccardo Tisci.
            
            \textbf{Retrieved Passages}: 

            [1] Title: Riccardo Tisci
            Text: Riccardo Tisci Riccardo Tisci (; born 1974) is an Italian fashion designer. He studied in Italy at the Design Istituto d’Arte Applicata in Cantù until the age of 17, and then graduated from London's Central Saint Martins Academy in 1999. In 2005, Tisci was given the title of creative director for Givenchy Women's haute couture and ready-to-wear lines. In May 2008 he was additionally named as menswear and accessories designer of the Givenchy men's division. In March 2018, it was announced he had been appointed chief creative officer of Burberry, succeeding Christopher Bailey. Tisci's apparent fascination with Gothic touches (dark,\\
            
            [2] Title: Riccardo Tisci
            Text: rappers Jay-Z and Kanye West to create the cover for "Watch the Throne", as well as the covers for singles H•A•M and Otis. He has now recently collaborated again with Kanye West for his label GOOD Music on their album "Cruel Summer". Riccardo Tisci Riccardo Tisci (; born 1974) is an Italian fashion designer. He studied in Italy at the Design Istituto d’Arte Applicata in Cantù until the age of 17, and then graduated from London's Central Saint Martins Academy in 1999. In 2005, Tisci was given the title of creative director for Givenchy Women's haute couture and ready-to-wear lines.\\
            
            [3] Title: Riccardo Tisci
            Text: Tisci worked for companies such as Puma and Coccapani before signing a three-year contract with Ruffo Research, a company that has helped launch the careers of several fashion designers, such as Sophia Kokosalaki. Tisci has mentioned in an interview that he would like to work with an Iranian brand, although it is not known how realistic his wish is. Upon the expiration of his contract in July 2004, Tisci spent time living in India, where he began to work on his own collection. In September 2004, during the Milan Fashion Week, Tisci debuted his first Riccardo Tisci Collection for Fall

             \noindent\rule{40em}{0.4pt}

            \textbf{Few-shot FT Baseline Response}: Several years after graduating from London's Central Saint Martins Academy in 1999, Riccardo Tisci was given the title of creative director for Givenchy Women's haute couture and ready-to-wear lines in 2005 [1]. Later, in May 2008, he was additionally named as menswear and accessories designer of the Givenchy men's division [1]. And in March 2018, he was appointed chief creative officer of Burberry, succeeding Christopher Bailey [3]. Tisci's apparent fascination with Gothic touches can be seen in his collaboration with rappers Jay-Z and Kanye West to create the cover for \"Watch the Throne\", as well as the covers for singles H•A•M and Otis [2]. He has also collaborated again with Kanye West for his label GOOD Music on their album \"Cruel Summer\" [2].\\

            \textbf{\projectname \ Response}: Riccardo Tisci is an Italian fashion designer, born in 1974 [1]. He studied at the Design Istituto d’Arte Applicata in Cantù in Italy until the age of 17 and later graduated from London's Central Saint Martins Academy in 1999 [1]. Tisci was appointed creative director for Givenchy Women's haute couture and ready-to-wear lines in 2005 [1]. In May 2008, he was also named as menswear and accessories designer for the Givenchy men's division [1]. In March 2018, Tisci was appointed chief creative officer of Burberry, succeeding Christopher Bailey [1]. Tisci has collaborated with rappers Jay-Z and Kanye West to create the cover for "Watch the Throne" and the covers for singles H•A•M and Otis [2]. He has also worked with Kanye West for his label GOOD Music on their album "Cruel Summer" [2].

            \end{minipage}}
        \caption{BIO Example of generated answers with citations given a question and retrieved passages. We compare the answers produced by the Few-shot FT baseline model with those generated by \projectname.}
        \label{app:bio-example}
\end{figure*}

\end{document}